\documentclass[pdflatex, referee, sn-nature]{sn-jnl}

\usepackage{amsmath}    % required by sn-jnl

\usepackage{booktabs}
\usepackage{siunitx}
\usepackage{graphicx}
\DeclareGraphicsExtensions{.pdf}

\begin{document}

% with primate prior and fine-tuning by human brain imaging features
\title{Neural Dynamics Model of Visual Decision-Making: Learning from Human Experts}
% \date{\today}

\author[1]{\fnm{Jie} \sur{SU}}
\equalcont{These authors contributed equally to this work.}
\author[1]{\fnm{Fang} \sur{CAI}}
\equalcont{These authors contributed equally to this work.}

\author[2]{\fnm{Shu-Kuo} \sur{ZHAO}}
\author[1]{\fnm{Xin-Yi} \sur{WANG}}

\author*[1]{\fnm{Tian-Yi} \sur{QIAN}}\email{qiantianyi@qiyuanlab.com}
\author*[2]{\fnm{Da-Hui} \sur{WANG}}\email{wangdh@bnu.edu.cn}
\author*[1]{\fnm{Bo} \sur{HONG}}\email{hongbo@qiyuanlab.com}

\affil*[1]{
    \orgname{Qiyuan Laboratory},
    \orgaddress{\city{Beijing}, \country{China}}%
    }
\affil[2]{
    \orgdiv{School of Systems Science},
    \orgname{Beijing Normal University},
    \orgaddress{
        \street{No. 19, Xinjiekouwai St.},
        \city{Beijing}, \postcode{100875}, \country{China}}%
    }

%%%%%%%%%%%%%%%%%%%%%%%%%%%%%%%%%%%%%%%%%%%%%%%%%
% \begin{abstract}
\abstract{
    Uncovering the fundamental neural correlates
    of biological intelligence,
    developing mathematical models,
    and conducting computational simulations
    are critical for advancing new paradigms
    in artificial intelligence (AI).
    In this study, we implemented
    a comprehensive visual decision-making model
    that spans from visual input to behavioral output,
    using a neural dynamics modeling approach.
    Drawing inspiration from the key components
    of the dorsal visual pathway in primates,
    our model not only aligns closely with human behavior
    but also reflects neural activities in primates,
    and achieving accuracy comparable to
    convolutional neural networks (CNNs).
    Moreover, magnetic resonance imaging (MRI)
    identified key neuroimaging features
    such as structural connections and functional connectivity
    that are associated with performance in perceptual decision-making tasks.
    A neuroimaging-informed fine-tuning approach
    was introduced and applied to the model,
    leading to performance improvements that
    paralleled the behavioral variations observed among subjects.
    Compared to classical deep learning models,
    our model more accurately replicates
    the behavioral performance of biological intelligence,
    relying on the structural characteristics
    of biological neural networks rather than extensive training data,
    and demonstrating enhanced resilience to perturbation.
}
% \end{abstract}

% \begin{keyword}
\keywords{
    neural dynamics model,
    spiking neural network,
    perceptual decision-making,
    magnetic resonance imaging,
    neuroimaging-informed fine-tuning
}
% \end{keyword}

\maketitle

%%%%%%%%%%%%%%%%%%%%%%%%%%%%%%%%%%%%%%%%%%%%%%%%%
\section{Introduction}
\label{sec:intro}

Abstracting the structures of biological neural systems
into mathematical models
and constructing artificial neural networks
based on these abstractions to address real-world problems,
represents a pivotal approach to innovation in artificial intelligence.
Over the past decades,
this approach has achieved significant success in various fields
such as pattern recognition and computer vision
\cite{mcculloch1943logical, hopfield1982neural,
lecun1998gradient, lecun2015deep, hassabis2017neuroscience}.
Visual perception, a fundamental process through which
humans and animals interpret and interact with the environment,
is a central topic in both neuroscience an artificial intelligence (AI).
Understanding the neural mechanisms underlying perceptual decision-making
not only provides insights into biological systems
but also has the potential to drive advancements in AI technologies.
Convolutional neural network (CNN) models,
inspired by the receptive fields and
parallel distributed processing in animal vision systems,
have achieved notable success in tasks such as
object detection, facial recognition
\cite{alex2012imagenet, szegedy2015going, he2016resnet},
and action recognition
\cite{ji2013conv3d, maturana2015voxnet},
often surpassing human performance.
However, these models face significant limitations,
including the need for vast amounts of labeled data,
lack of biological interpretability,
and susceptibility to adversarial attacks.
These limitations highlight the gap between
current AI systems and biological neural networks
regarding robustness, efficiency, and flexibility.
Addressing these challenges requires a paradigm shift
towards models that more closely mimic the
underlying principles of biological neural computation.

Motion perception is critical for animals
to detect potential conspecifics, preys or predators,
and is vital for their survival.
The motion perception system starts with retinal input,
travels through the lateral geniculate nucleus (LGN)
to the primary visual cortex (V1),
and then projects to parietal areas along the dorsal visual pathway,
supporting spatial attention and eye movements
\cite{mishkin1982contribution}.
Direction-selective (DS) neurons
are widely present in brain areas such as V1,
middle temporal area (MT),
and lateral intraparietal area (LIP)
\cite{shadlen2001neural, katz2016dissociated, yates2017functional},
which are essential for motion perception.
The direction preference of V1 DS neurons depends on
the spatial pattern of LGN neurons to which they are connected
\cite{chariker2021theory, chariker2022computational}.
Motion perception relies on the integration of
both spatial and temporal dimensions of information
\cite{burr2011motion}.
Spatial integration is achieved
by the larger receptive fields of MT neurons
\cite{amano2012human, nishida2018motion},
while temporal integration primarily relies on
the temporal integration properties of LIP neurons
\cite{shadlen1996motion, shadlen2001neural}.

Based on the physiological structures and properties
of neurons and circuits in the LIP area,
Wang proposed a recurrent neural circuit model
\cite{wang2002probabilistic}.
Within this model,
two groups of direction-selective excitatory neurons
are capable of integrating lower-level synaptic inputs
and performing cognitive tasks through
recurrent excitation and mutual inhibition mechanisms.
The model's attractor dynamics enhance its ability
to perform decision-making and working memory tasks,
replicate neural activity patterns observed in non-human primates,
and achieve performance consistent with experimental data.
Subsequent researches have extended the model by focusing on
its simplification, theoretical analyses and training methodologies
\cite{wong2006recurrent, wang2008decision, wang2020macroscopic, song2017reward}.
These studies primarily emphasize
fitting and interpreting biological experimental data,
showcasing significant biological plausibility and interpretability.
However, neuron parameters are usually determined
based on averaged experimental data,
without considering the physiological characteristics
that underlie behavioral differences among individuals.
Moreover, the model has not been optimized
based on biological features found in primates,
which limits its impact and application in artificial intelligence.

Numerous studies have investigated the relationship
between the physiological or structural features
of the human brain and behavior
% (termed structural brain-behavior association)
\cite{genon2022linking}.
These studies span various human behaviors,
including cognitive functions such as
working memory \cite{colom2007general},
language acquisition \cite{hamalainen2017bilingualism},
theory of mind \cite{rice2015spontaneous},
and social functions like
social network size \cite{kanai2012online}.
Some research examines differences in
physiological characteristics between healthy and diseased groups,
for example, correlations between neurodegenerative diseases
and physiological indicators of white matter fiber tracts
\cite{forkel2022white, schmithorst2010white, kantarci2017white}.
Modern neuroimaging technologies provide effective tools
for quantifying human brain physiological features, including
gray matter volume \cite{mechelli2005voxel},
cortical thickness \cite{fischl2000measuring},
and myelination \cite{glasser2011mapping}
calculated from structural images,
measurements of white matter areas from diffusion images
\cite{pievani2010assessment},
and the correlation of blood-oxygen-level-dependent (BOLD) signals
between brain areas in resting state,
known as resting-state functional connectivity (rest FC)
\cite{sui2014function, wang2007altered, song2008brain}.
These studies have revealed correlations between
brain structure, functional characteristics, and human behavior,
contributing to understanding disease mechanisms
\cite{hashemi2020bayesian}.
Despite these insights,
neuroimaging data has not been fully integrated into
the development or optimization of artificial neural network (ANN) models.
Integrating neuroimaging data with ANN models
offers a promising frontier for creating brain-inspired AI models
that replicates the behavioral performance of human subjects.
% that more accurately reflect the complexities of biological neural systems.

Addressing the gap in biologically inspired dynamics modeling
in artificial intelligence applications,
specifically the lack of tuning mechanisms
integrated with behavioral and neuroimaging data,
this study synthesizes known neural mechanisms
to construct a comprehensive biological neural dynamics model
of visual motion perception.
The model adheres to the physiological features of non-human primates
and facilitates cognitive decision-making behavior
in the Random Dot Kinematogram (RDK) task,
exhibiting biological-like behavioral and neuronal characteristics.
Furthermore, this study explores the correlation between
behavioral and neuroimaging features of human subjects
during visual decision tasks through
magnetic resonance imaging (MRI) techniques,
particularly focusing on the structural and functional characteristics
of corresponding brain regions in human experts.
Finally, we introduce a novel
\emph{neuroimaging-informed fine-tuning} approach,
which leverages these neuroimaging characteristics
to optimize the artificial biological neural network model
and achieve notable improvements in performance.

%%%%%%%%%%%%%%%%%%%%%%%%%%%%%%%%%%%%%%%%%%%%%%%%%
\section{Results}
\label{sec:results}

\subsection{A Neural Dynamics Model of Visual Decision-Making with
Behavioral Performance and Neural Activities Similar to Primates}
\label{result:active}

\begin{figure}[htb!]
    \includegraphics[width=\linewidth]{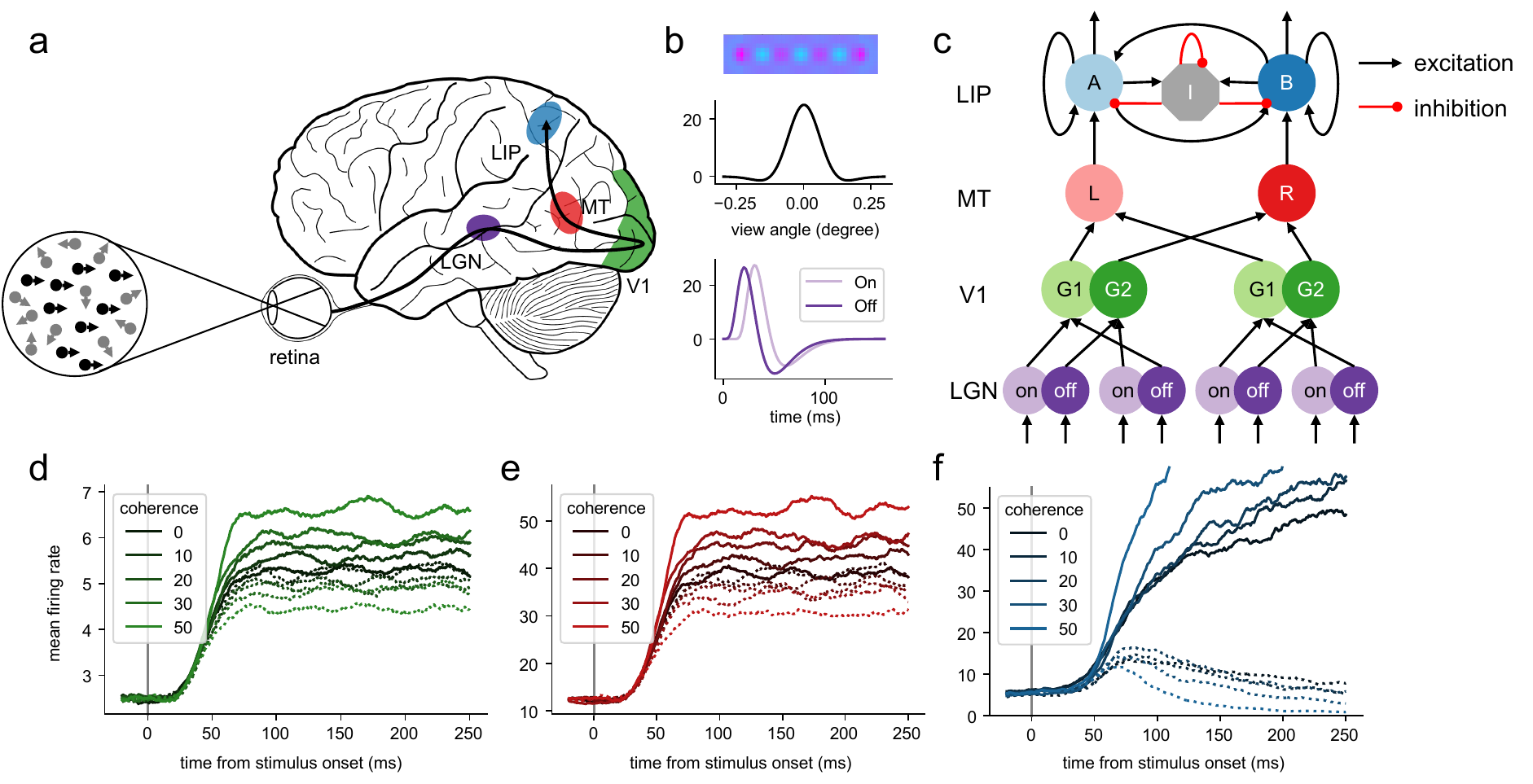}
    \caption[A neural dynamics model of motion perception
            inspired by the dorsal visual pathway]{
        \label{fig:model}
        \emph{A neural dynamics model of motion perception
            inspired by the dorsal visual pathway.}
    % }
    % \justifying
    % \begin{enumerate*}[label=\textbf{\alph*.}]
        % \item
        \textbf{a}.
        The Random Dot Kinematogram (RDK)
        and the locations of the key structures within the dorsal visual pathway
        associated with motion perception and decision-making in the human brain.
        % \item
        \textbf{b}.
        Spatial characteristics and arrangements of LGN \emph{ON} and
        \emph{OFF} neurons, including their spatial and temporal profiles.
        % \item
        \textbf{c}.
        Structure of the biologically inspired neural dynamics model,
        illustrating four modules corresponding to those key brain regions
        and the connections between them.
        % \item
        \textbf{d-f}.
        Averaged firing rates of distinct groups of neurons
        in response to the RDK stimulus.
        Time 0 marks the stimulus onset.
        The solid line represents neuron groups with a preferred directional bias,
        while the dotted line represents neuron groups with an opposite bias.
        Color and legend indicate coherence levels.
        \textbf{d}. V1 neurons demonstrate direction selectivity;
        \textbf{e}. MT neurons enhance this selectivity, with the activation
            of direction-preference neurons increasing with coherence;
        \textbf{f}. LIP neurons exhibit the ramping activity and a
            \emph{winner-take-all} effect,
            with the slope of ramping increasing as coherence levels rise.
    % \end{enumerate*}
    }
\end{figure}

We constructed a neural dynamics model that
encompasses the four key brain areas of the dorsal visual pathway
(LGN, V1, MT, and LIP, Figs.~\ref{fig:model}a-c).
This model forms an artificial biological neural network
capable of motion perception and performing
the Random Dot Kinematogram (RDK) task.
The neurons in our model adopt the Leaky Integrate-and-Fire (LIF) model
and are interconnected through excitatory synapses (AMPA, NMDA)
and inhibitory synapses (GABA) (see \S~\ref{method:model}).
The spikes of different neuron groups were recorded
during the RDK task (Fig.~\ref{fig:model}d-f).

The electrophysiologic recordings show that
V1 neurons exhibit direction selectivity,
which is further enhanced in the MT neurons.
In the LIP area,
neurons that prefer a specific motion direction gradually dominate,
while those favoring the opposite direction are suppressed,
resulting in a ``\emph{winner-take-all}'' effect.
Additionally, V1 and MT neuron activation increases with motion coherence
but stabilizes during the stimulus period.
In contrast, LIP neurons show a gradual ramping of activation,
with the ramping speed proportional to motion coherence
(Fig.~\ref{fig:model}d-f).
These simulated neural activities closely resemble those
recorded in electrophysiological experiments with macaque monkeys
\cite{hanks2006microstimulation, katz2016dissociated},
demonstrating the biological plausibility of our model.

\begin{figure}[htb!]
    \includegraphics[width=\linewidth]{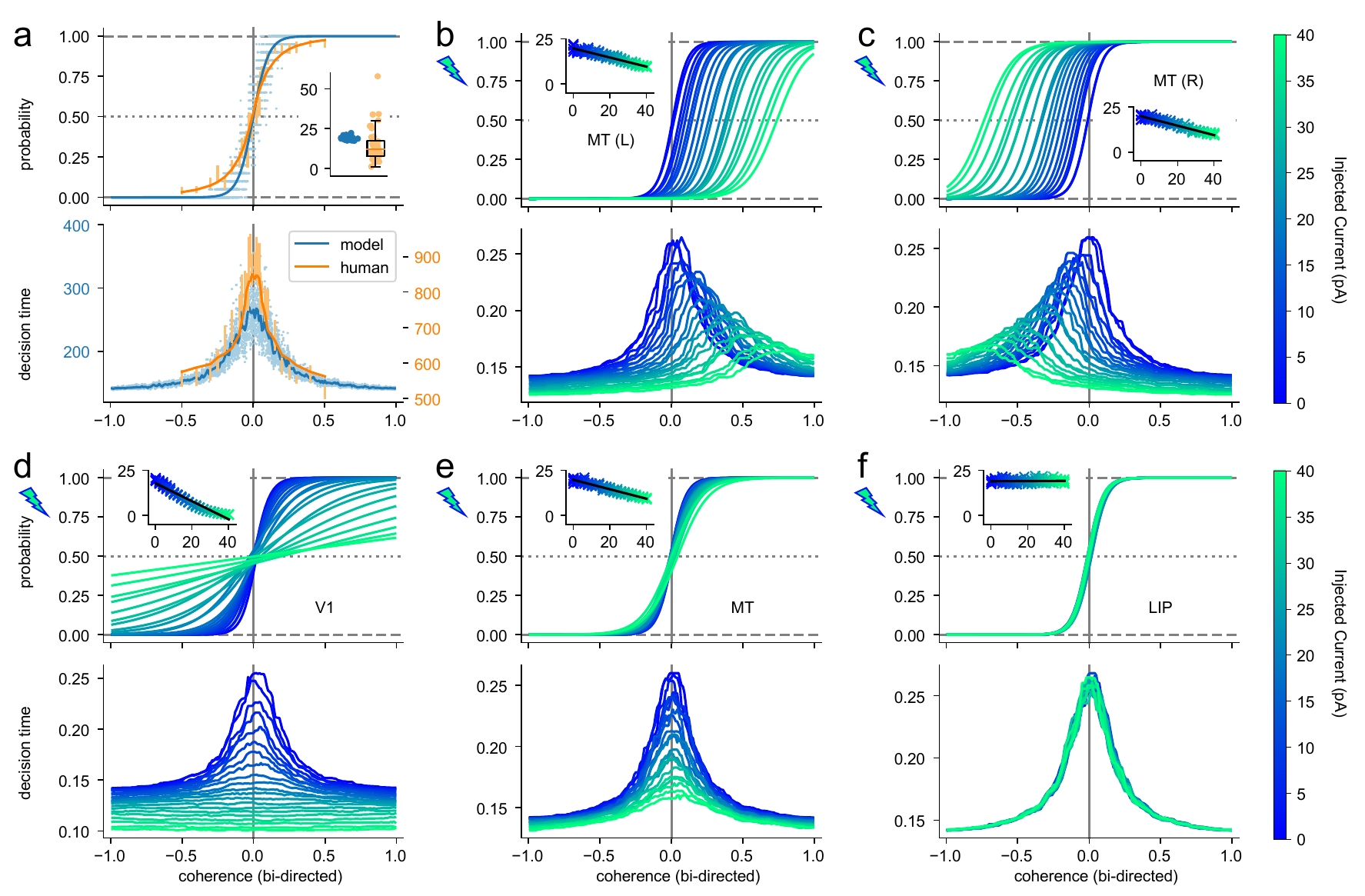}
    \caption[Behavioral performance of the model in RDK tasks
            and effect of virtual electrical stimulation]{
        \label{fig:behavior}
        \emph{Behavioral performance of the model in RDK tasks
            and effect of virtual electrical stimulation.}
    % }
    % \justifying
    % \begin{enumerate*}[label=\textbf{\alph*.}]
        % \item
        \textbf{a}.
        Behavioral performance of the model and human subjects in the RDK task,
        showing selection probability, and average reaction time
        as functions of coherence level
        (upper: psychometric curve, lower: decision time curve).
        The model's performance is represented in blue (average of 45 repetitions),
        and human performance is shown in orange (average of 36 subjects).
        Error bars denote the standard error of the mean (SEM) across all subjects.
        The inset subplot displays the distribution of sensitivity ($k$)
        for all subjects (orange) and the model (blue, repeated 45 times).
        % \item
        \textbf{b}.
        When additional current (\qtyrange{0}{40}{\pA}) is injected into
        the `\emph{L}' group of neurons in the MT area of the model,
        the psychometric and decision time curves shift rightward,
        sensitivity decreases (slope\num{=-0.25}, $p<0.001$),
        and reaction time shorten.
        Color and colorbar represent the intensity of stimulation,
        consistent throughout.
        % \item
        \textbf{c}.
        Injecting additional current into the `\emph{R}' group
        of neurons in MT results in a leftward shift
        in the psychometric and decision time curves,
        with decreased sensitivity and decision time
        (slope\num{=-0.26}, $p<0.001$).
        % \item
        \textbf{d}.
        Performance changes when additional electrical stimulation
        is applied to all neurons in the model's V1.
        The additional input current decreases
        the slope of the psychometric curve
        (indicating reduced sensitivity, slope\num{=-0.50}, $p<0.001$)
        and reduces mean decision time.
        % \item
        \textbf{e}.
        Sensitivity and decision time both decrease when
        additional electrical stimulation is applied to all neurons in MT
        (slope\num{=-0.26}, $p<0.001$).
        % \item
        \textbf{f}.
        Applying the same intensity of current to LIP
        has negligible effects on the model's performance
        (slope\num{=0.005}, $p=0.60$).
    % \end{enumerate*}
    }
\end{figure}

The performance of our model was evaluated using the entire
Random Dot Kinematogram (RDK) dataset (see \S~\ref{method:rdk}).
We calculated the choice probability
and average reaction time for each coherence level.
Fig.~\ref{fig:behavior}a (upper panel)
presents the psychometric curve of the model (in blue),
which is similar to the psychometric curves of human subjects (in orange).
The model's sensitivity
(slope of the psychometric curve, $k$ in Equation~\ref{eq:psychometric})
is $19.31\pm 0.17$ (mean\textpm SEM of 45 experiments),
significantly exceeding the average level of human subjects
(36 subjects, $15.1\pm 1.9$, $t=2.50$, $p=0.015$, two-sample $t$-test).
Moreover, the model's decision time curve
aligns with the reaction times of human subjects.
As coherence increases (task difficulty decreases),
the decision time gradually becomes shorter,
as illustrated in Fig.~\ref{fig:behavior}a (lower panel).

In electrophysiological experiments,
microstimulation of specific brain regions
can induce behavioral changes
\cite{hanks2006microstimulation, katz2016dissociated}.
To verify whether our neural dynamics model
exhibits similar characteristics to biological motion perception systems,
we conducted virtual cortical stimulation experiments.
External currents were injected
into different groups of neurons in our model
to investigate if the results matched those observed in biological experiments
\cite{hanks2006microstimulation, katz2016dissociated}.

Applying continuous extra current
(\qtyrange{0}{40}{\pA}, Fig.~\ref{fig:behavior}b)
to MT neurons that prefer leftward motion
shifted the model's psychometric curve to the right,
indicating an increased leftward choice preference
(slope\num{=-0.18}, $p<0.001$),
% slope=-0.1788162107265768, intercept=-0.4057070813286381,
% rvalue=-0.9814195367607182, pvalue=1.130194209467442e-75
%%
This manipulation slightly reduced the model's sensitivity
(slope\num{=-0.25}, $p<0.001$),
% slope=-0.2518379433780535, intercept=19.767667460124308,
% rvalue=-0.9616052024904997, pvalue=1.1766832852340617e-59
and altered decision times,
decreasing for leftward and increasing for rightward motions,
aligning with observations in animal studies
\cite{hanks2006microstimulation}.
Conversely, applying extra current to neurons
that prefer rightward motion resulted in
a leftward shift of the psychometric curve
and a corresponding leftward shift in the decision time curve
(Fig.~\ref{fig:behavior}c,
intercept changed with external currents: slope\num{=0.18}, $p<0.001$;
% slope=0.17806665033557414, intercept=0.439576365240296,
% rvalue=0.9811459156144616, pvalue=2.38315214543158e-75
sensitivity changes with external currents: slope\num{=-0.26}, $p<0.001$).
% slope=-0.2626473844910249, intercept=20.009859642387646,
% rvalue=-0.9717810715087215, pvalue=1.9711286889678835e-66
%%
Moreover, when stimulating both groups of neurons
in the MT region without selection
(\qtyrange{0}{40}{\pA}),
the model's sensitivity slightly declined
(slope\num{=-0.26}, $p<0.001$),
% slope=-0.2639976559921203, intercept=19.789140247675192,
% rvalue=-0.9754273508302072, pvalue=1.737334140075454e-69
but the decision time notably reduced
(Fig.~\ref{fig:behavior}e).
Correspondingly, stimulating all neurons in the V1 region
significantly altered the model's sensitivity
(slope\num{=-0.50}, $p<0.001$)
and decision time (Fig.~\ref{fig:behavior}d).
% slope=-0.49987795087271625, intercept=17.779940981067323,
% rvalue=-0.9754162846364444, pvalue=1.7775962861143678e-69
%
However, stimulating all excitatory neurons in the LIP area
did not affect the model's sensitivity
(slope=\num{0.005}, $p=0.60$)
% slope=0.00451279253063507, intercept=18.98739724427212,
% rvalue=0.05202695280718643, pvalue=0.5981289094522815
or decision time (Fig.~\ref{fig:behavior}f).

\subsection{Model Optimization Inspired by Structural Connectivity of Human Experts}
\label{result:connect}

Structural connectivity estimated from neuroimaging techniques
reflects the mesoscopic properties of fibers,
which are directly related to behavioral outcomes
\cite{genon2022linking, colom2007general,
hamalainen2017bilingualism, rice2015spontaneous, kanai2012online}.
To identify the key parameters
influencing the task performance,
we collected behavioral and neuroimaging data
from 36 human subjects performing the RDK tasks.

\begin{figure}[htb!]
    \includegraphics[width=\linewidth]{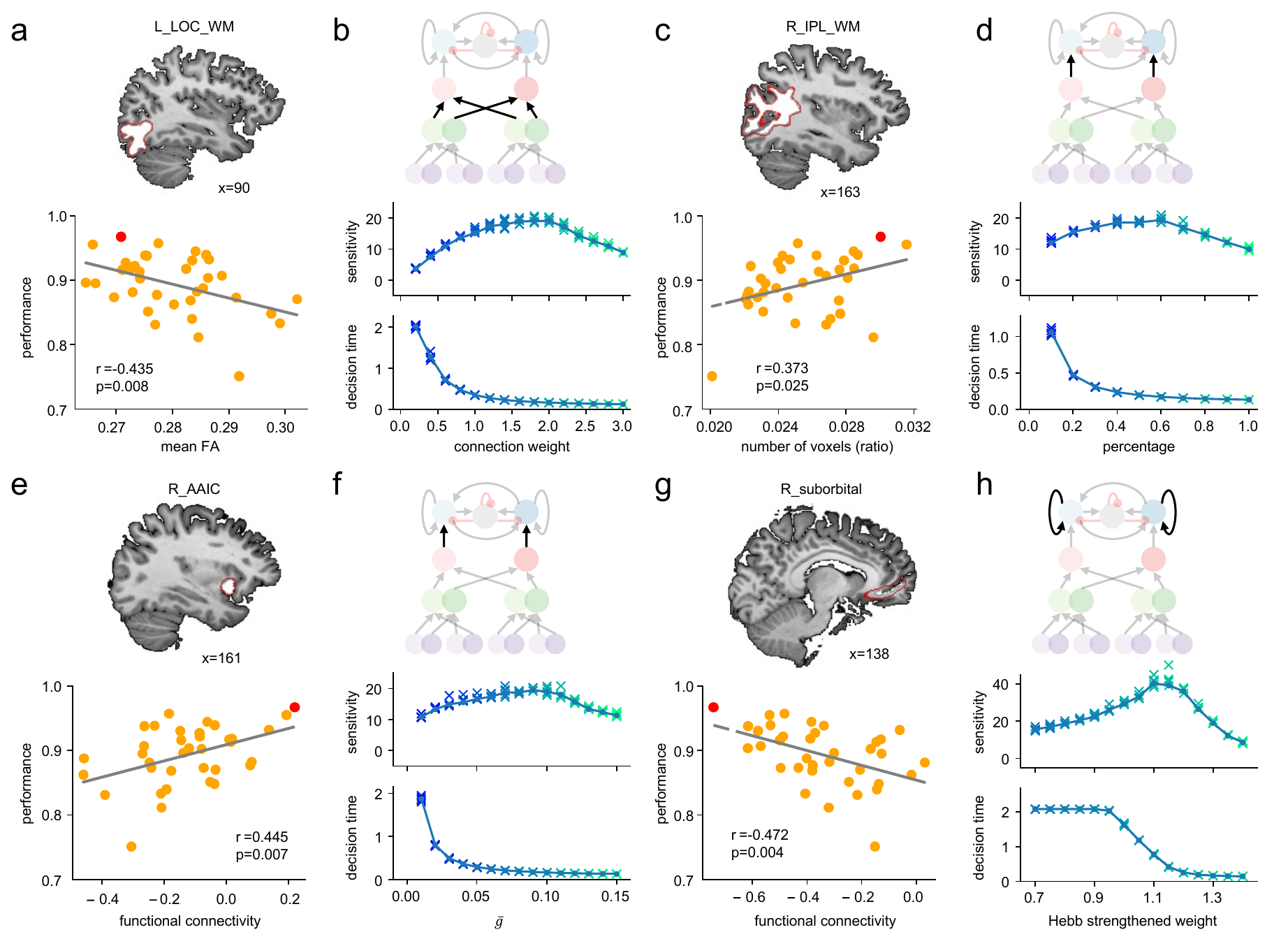}
    \caption[Neuroimaging-informed model tuning]{
        \label{fig:tuning}
        \emph{Neuroimaging-informed model tuning.}
    % }
    % \justifying
    % \begin{enumerate*}[label=\textbf{\alph*.}]
        % \item
        \textbf{a}.
        Mean FA values of the left lateral occipital area in V1
        % (upper, L-LOC-WM, marked as the white ROI with a red contour line)
        were negatively correlated with subjects' task performance.
        % ($r=-0.435$, $p=0.008$, uncorrected Pearson correlation).
        The red dot represents the best-performing subject
        (hereinafter the same).
        Subjects who exhibited lower anisotropy (higher fiber bundle density)
        in this region performed better in the task.
        % \item
        \textbf{b}.
        Tuning the average connection weight between V1 and MT in the model
        improved accuracy and efficiency within a specific range.
        Upper: Schematic diagram of the adjusted V1--MT connection.
        Lower: Increasing the connection weight initially
        improves model sensitivity and reduces decision time,
        but later stages show a decrease in sensitivity
        while decision time continues to decrease.
        % \item
        \textbf{c}.
        The white matter volume in the right inferior parietal lobe,
        % (R\_IPL\_WM, upper, marked as the white ROI with a red contour line),
        located between the occipital and parietal lobes,
        was positively correlated with task performance.
        % ($r=0.373$, $p=0.025$, uncorrected Pearson correlation).
        This indicates that subjects with better performance
        had larger white matter volume relative to total brain volume.
        % \item
        \textbf{d}.
        Tuning the proportion of connections between MT and LIP
        within a specific range improved the model's accuracy and efficiency.
        Upper: Schematic diagram of the adjusted MT--LIP connection.
        Lower: Increasing the connection ratio
        initially enhances sensitivity and reduces decision time,
        but excessive connections can decrease sensitivity.
        % \item
        \textbf{e}.
        Resting-state functional connectivity between
        MT and AAIC in the right hemisphere
        % (R-AAIC, upper, marked as the white ROI with a red contour line)
        was positively correlated with task performance.
        % ($r=0.445$. $p=0.007$, uncorrected Pearson correlation). 
        The positive connectivity may reflect enhanced
        self-monitoring during the task, leading to improved performance.
        % \item
        \textbf{f}.
        Increasing the average synaptic conductivity in MT within
        a specific range improved the model's sensitivity and efficiency.
        Upper: Schematic diagram of the adjusted MT--LIP connections.
        Lower: As conductivity increases,
        sensitivity initially rises and decision time decreases,
        but sensitivity eventually declines
        despite continued reduction in decision time.
        % \item
        \textbf{g}.
        Functional connectivity between
        LIP and the suborbital region in the left hemisphere
        % (R-suborbital, upper, marked as the white ROI with a red contour line)
        was negatively correlated with task performance.
        % ($r=-0.472$, $p=0.004$, uncorrected Pearson correlation). 
        The negative connectivity may reflect
        increased engagement during the task,
        which could lead to improved performance.
        % \item
        \textbf{h}.
        Increasing the \emph{Hebb-strengthened weight}
        between LIP excitatory neurons within a specific range
        improved the model's sensitivity and efficiency.
        Upper: Schematic diagram of the adjusted LIP recurrent connections.
        Lower: Increasing the weight
        initially enhances sensitivity and reduces decision time,
        but excessive weight can decrease sensitivity.
        The Pearson correlation (uncorrected) was used for these analyses;
        $p$-values are reported.
    % \end{enumerate*}
    }
\end{figure}

In the correlation analysis,
we found a significant negative correlation between
the mean fractional anisotropy (FA) value
% (a structural indicator estimated from the diffusion MR image)
of white matter in the left lateral occipital region
and subjects' behavioral performance
($r=-0.435$, $p=0.008$, uncorrected Pearson correlation,
Fig.~\ref{fig:tuning}a).
This indicates that
subjects with lower mean FA values in this area
performed better in the behavioral experiment.
The lateral occipital region, located in the occipital lobe,
includes white matter pathways connecting
the primary visual cortex and the MT area.
Lower FA values may reflect reduced anisotropy
and predict a denser distribution of fiber bundles in this region.

The relative white matter volume in the right inferior parietal region
(i.e., the ratio of ROI volume to total brain volume,
estimated from MR T1 images)
was also significantly positively correlated
with subjects' behavioral performance
($r = 0.373$, $p = 0.025$, uncorrected Pearson correlation,
Fig.~\ref{fig:tuning}c).
This suggests that subjects with better performance
had a larger white matter volume in this area,
indicating a broader range of fiber tracts.
The inferior parietal region,
located between the occipital and parietal lobes,
includes the white matter pathway connecting the MT and LIP regions,
which might be associated with decision-making in the brain.
These results demonstrate correlations between behavioral performance
and the structural features of the visual dorsal pathway.
No other structural features showed significant correlations.

In response to key parameters identified in MRI studies,
specifically the white matter connections
from V1 to MT and from MT to LIP,
which influence behavioral performance,
we employed a \emph{neuroimaging-informed fine-tuning} approach
to adjust model parameters based on MRI analysis results.
For the mean FA of left lateral occipital area
(Fig.~\ref{fig:tuning}a),
we modified the average connection weight
of the V1-to-MT connections in the model
(Fig.~\ref{fig:tuning}b, upper panel) accordingly.
Increasing the connection weight at lower strengths
(mean of the connection matrix gradually increasing to 2)
enhanced model sensitivity and reduced decision times,
aligning with the observed behavioral correlations.
However, further increases in the connection weight
led to reduced decision times but a decline in sensitivity
(Fig.~\ref{fig:tuning}b, lower panel).
Similarly, adjusting the connection ratio between MT and LIP neurons
(Fig.~\ref{fig:tuning}d upper panel)
revealed that increasing the ratio from low levels
improved model sensitivity and reduced decision times,
aligning with neuroimaging and behavioral correlations.
However, beyond a certain range,
further increases in the connection ratio
decreased both model sensitivity and decision times
(Fig.~\ref{fig:tuning}d lower panel).

\subsection{Model Optimization Inspired by Functional Connectivity of Human Experts}
\label{result:functional}

% We also analyzed the correlation between
% functional connectivity indicators
% and behavioral performance in human subjects.
The correlation between the functional connectivity indicators
and the behavioral performance in human subjects was also analyzed.
We found a significant positive correlation between
behavioral performance and resting-state functional connectivity (FC)
of the MT region and the anterior agranular insular complex (AAIC.
$r=0.445$, $p=0.007$, uncorrected Pearson correlation,
Fig.~\ref{fig:tuning}e).
The resting-state FC between the two brain regions
was more positively correlated in subjects with better performance,
which could indicate enhanced self-monitoring during the task.

The resting-state FC between the LIP region
and the suborbital region of the prefrontal lobe
was significantly negatively correlated with behavioral performance
($r=-0.472$, $p=0.004$, uncorrected Pearson correlation,
Fig.~\ref{fig:tuning}g).
This suggests that a stronger negative correlation
between resting-state activities in these regions
is associated with better performance,
possibly reflecting higher engagement or attentiveness during the task.
No significant correlations were observed
between the resting-state FC features of other regions and subject behavior.

The functional MRI results highlighted brain areas
not included in the existing dynamics model.
These regions may provide top-down control
to areas involved in motion perception and decision-making,
thereby affecting perceptual decisions and behavior.
Based on findings from electrophysiological experiments
\cite{briggs2013attention},
we simulated the modulation of the MT region
by adjusting synaptic conductance in this module,
without adding new brain regions to the model.
The results are illustrated in Figs.~\ref{fig:tuning}f and h.
Increasing MT synaptic conductance enhanced
model sensitivity and reduced decision time,
aligning with observed subject behavior.
However, beyond a certain point,
further increases in connection strength
led to decreased sensitivity despite reduced decision time.
Similarly, we simulated the regulation of the LIP region
by adjusting connection efficiency between DS neurons.
This simulation, detailed in Fig.\ref{fig:tuning}h,
showed that increasing connection efficiency
improved model performance and reduced decision time,
but excessive efficiency negatively affected performance.

These experiments illustrate how adjusting model parameters
based on structural and functional features
can enhance the neural dynamics model,
reflecting biological behaviors observed in neuroimaging studies.
This approach not only validates the model's biological plausibility
but also introduces a novel method for incorporating neuroimaging data
into the fine-tuning of artificial neural networks,
which we term \emph{neuroimaging-informed fine-tuning}.

\subsection{Superior Robustness of Neural Dynamics Model Under Perturbation}
\label{result:noise}

\begin{figure}[ht!]
    \includegraphics[width=\linewidth]{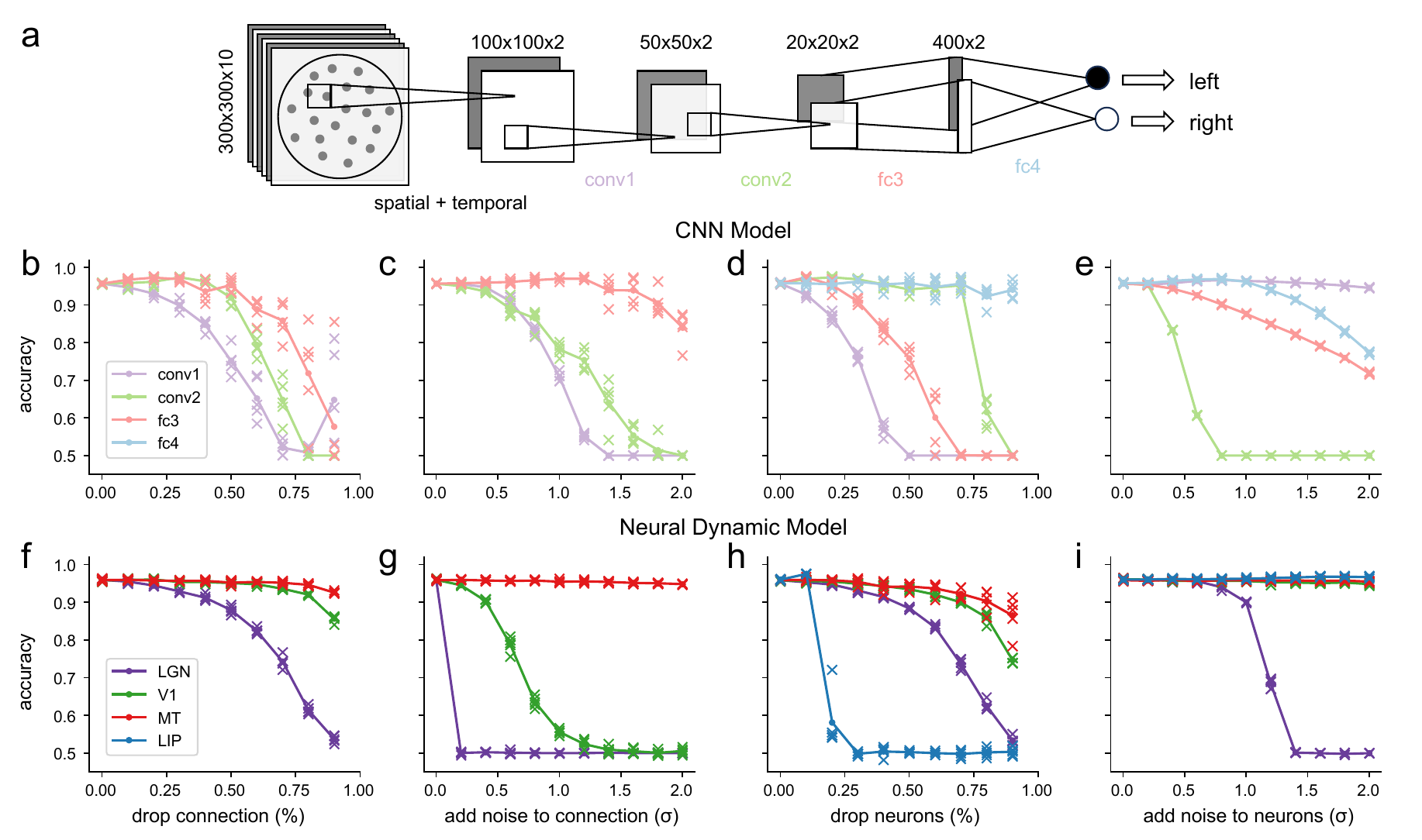}
    \caption[Comparison between CNN and the neural dynamics model in terms of perturbation]{
        \label{fig:noise}
        \emph{Comparison between CNN and the neural dynamics model in terms of perturbation.}
    % }
    % \justifying
    % \begin{enumerate*}[label=\textbf{\alph*.}]
        % \item
        \textbf{a}.
        The structure and parameters of the CNN model.
        % \item
        \textbf{b-e}.
        In the CNN model, changes in accuracy
        under different perturbation conditions:
        \textbf{b}.
        when a certain percentage of connections
        are dropped in each layer;
        \textbf{c}.
        when zero-mean Gaussian noise
        (with variance as a multiplier of the average absolute value)
        is added to the connection weights;
        \textbf{d}.
        when a certain percentage of neurons are dropped;
        and
        \textbf{e}.
        when zero-mean Gaussian noise
        is added to the neuron activations.
        The colors represent the various parameter layers of the network
        (b, c, excluding the fully connected layer) or the presynaptic neurons (d, e).
        Cross marks indicate the accuracy of individual experiments,
        while the curves represent the overall accuracy
        averaged across five experiments.
        % \item
        \textbf{f-i}.
        In the neural dynamics model,
        changes in accuracy under similar perturbation conditions:
        \textbf{f}.
        when a certain percentage of connections are dropped in each layer;
        \textbf{g}.
        when zero-mean Gaussian noise
        is added to the connection weights;
        \textbf{h}.
        when a certain percentage of neurons are dropped;
        and
        \textbf{i}.
        when zero-mean Gaussian noise
        is added to the neuron activations.
        The colors represent the neurons in each layer (h, i)
        or the connections emanating from each layer (f, g).
    % \end{enumerate*}
    }
\end{figure}

To assess the robustness of our model,
we conducted four types of perturbation experiments
on both the neural dynamics model
and the convolutional neural network model,
as described in Section~\ref{method:noise}.
Noise was introduced by either
discarding or adding noise to the connection weights
or neurons in each module of the models.
Figs.~\ref{fig:noise}b--i
show the changes in accuracy
of the CNN and neural dynamics models at each layer
with varying perturbation intensities
(see Table~\ref{tab:compare} for statistics,
and Extended Fig.~\ref{fig:slope} for changes in sensitivity).

From these figures,
it is evident that our neural dynamics model
performs better compared to the CNN model under perturbation.
The CNN's accuracy typically declines
more significantly with increased perturbation,
whereas the neural dynamics model remains more stable overall.
Notably, the model's performance remains largely unaffected
when noise is introduced into modules
corresponding to higher-level brain regions.
For example, discarding connections from MT to LIP
(Fig.~\ref{fig:noise}f, red)
or adding noise to these connections
(Fig.~\ref{fig:noise}g, red)
has minimal impact on model accuracy,
far less than the perturbation effects on the CNN
(Figs.~\ref{fig:noise}b\&c, red).
Adding noise to the input current of neurons
almost does not affect the neural dynamics model's performance
(Fig.~\ref{fig:noise}i, except the purple).
Adding noise to LGN neurons causes model failure
(Figs.~\ref{fig:noise}g\&i, purple),
likely due to the sparse connections between V1 and LGN
leading to over-reliance on LGN signals.
Discarding neurons in the decision layer LIP
(Fig.~\ref{fig:noise}g, blue)
also drastically reduces model performance,
as too few LIP neurons cannot maintain
the attractor states corresponding to decisions,
causing the model to rapidly degrade to a resting state
with a single attractor \cite{ye2021quantifying},
effectively rendering it unable to make decisions.

Additionally, we found that changes in accuracy and sensitivity
with added noise in the neural dynamics model are consistent,
which contrasts with the behavior observed in CNN
In some cases, the CNN model retains high sensitivity,
but its accuracy significantly decreases with increasing model bias.
Furthermore, the CNN model exhibits greater variation with perturbation
(see scatter distribution in Figs.~\ref{fig:noise}b--d,
see also Extended Figs.~\ref{fig:another}, \ref{fig:onemore}),
while the neural dynamics model shows smaller bias
and maintains relatively stable performance even when disturbed.
Unexpectedly, reducing the number of LIP neurons
(Extended Fig.~\ref{fig:slope}g, blue)
and adding noise to LIP neuron input currents
(Extended Fig.~\ref{fig:slope}h, blue)
may improve model performance.
This is because our base model is not fully optimized;
weakening LIP neuron recurrent connections
or increasing Ornstein--Uhlenbeck noise may improve performance,
thus similar changes in perturbation
may also enhance model performance.

%%%%%%%%%%%%%%%%%%%%%%%%%%%%%%%%%%%%%%%%%%%%%%%%%
\section{Discussion}
\label{sec:discussion}

%%%% overall
In this study, we developed a neural dynamics model
for motion perception and decision-making
that simulates the visual dorsal pathway.
The model incorporates biomimetic neurons and synapses
with dynamic characteristics and kinetic properties as its core elements.
These components enable the model to exhibit
neural activities and behavioral outputs
that are comparable to those of the biological brain,
including responses to virtual electrical stimulation.
By leveraging neuroimaging insights
such as the structural features identified from human experts,
we employed neuroimaging-informed fine-tuning
to optimize the model parameters,
leading to improved performance.
Compared to CNNs, our model achieves
similar performance with fewer parameters,
more closely aligns with biological data,
and exhibits greater robustness to perturbations.

%%%% tuning: question and what was done
Parameter optimization for neural dynamics model
is both critical and challenging due to
the complexity of nonlinear dynamic systems.
Previous studies have primarily employed
data fitting techniques using neural or behavioral data,
which are feasible for small-scale models.
However, as neural dynamics models increase in scale and complexity,
with larger number of parameters and higher computational demands,
optimization become significantly more computationally intensive and difficult.
while it is theoretically possible to optimize
key parameters in some simplified models through analytical methods,
such method is generally impractical for more complex models.
This study, for the first time,
explores the application of neuroimaging data in model optimization,
a process we term \emph{neuroimaging-informed fine-tuning}.
While this approach may not achieve the global optimum,
it offers practical directions for parameter optimization
and significantly reduces the search space.
It is noteworthy that excessive adjustments to
parameters can impair model performance.
This observation implies the existence of an optimal range
for these parameters that maximizes performance.
Biological neural systems,
refined over billions of years of evolution,
regulate these parameters within such an optimal range,
with minor variations accounting for individual differences.

%%%% tuning: how
Targeting neuroimaging-informed fine-tuning,
we identified brain regions associated with
visual decision-making in human subjects
through behavioral and imaging experiments.
Subjects exhibiting superior performance in visual tasks
tend to have lower mean FA values in the occipital region,
indicating a higher degree of neural branching in the lateral occipital area.
Variations in white matter characteristics within the adults
are thought to result from inherent brain structure
and the long-term maturation process from childhood to adulthood
\cite{lebel2008microstructural},
suggesting that individuals with advanced visual capabilities
likely have undergone extensive training and development in visual function.
We also identified correlations between the functional connectivity of
MT--AAIC and LIP--suborbital with behavioral performance.
Variations in functional connectivity between subjects
are primarily influenced by their level of participation
and cognitive engagement during experiments \cite{kuhn2021brain},
suggesting that expert subjects are more engaged and attentive during tasks.
We hypothesize that this process involves top-down regulation
from higher-level brain areas to task-related regions,
potentially enhancing synaptic transmission efficiency.
The structural and functional connectivity features
observed in human subjects account for their superior performance
from the perspectives of innate structure,
long-term development, and short-term participation.
These findings, applied to key parameter adjustments in the model,
indicate that the neural dynamics model aligns with biological intelligence
and shows good interpretability in physiological terms.
This proposes a potential path for enhancing the model's performance
through neuroimaging-informed fine-tuning.

\begin{figure}[ht!]
    \includegraphics[width=\linewidth]{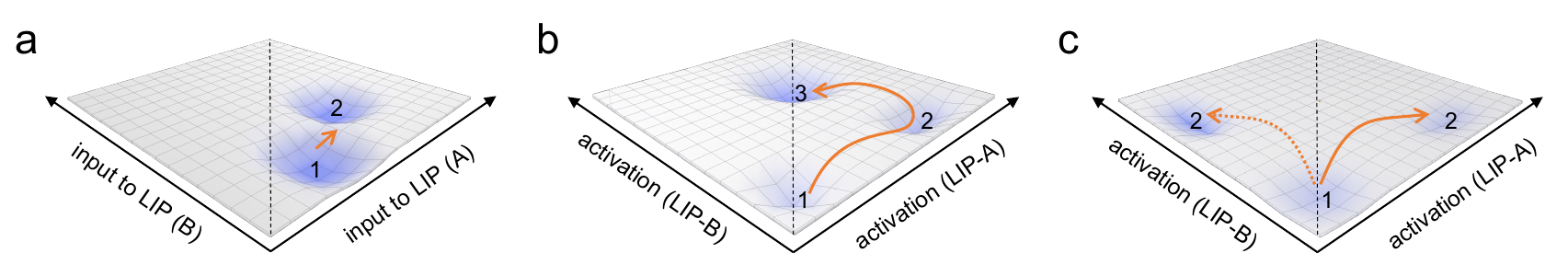}
    \caption[Diagram showing the impact of parameter adjustments on the model]{
        \label{fig:diagram}
        \emph{Diagram showing the impact of parameter adjustments on the model.}
    % }
    % \justifying
    % \begin{enumerate*}[label=\textbf{\alph*.}]
        % \item
        \textbf{a}.
        Adjusting connections between V1 and MT or MT and LIP
        alters the information received by LIP.
        Increasing connection weights shifts the input currents
        received by the two groups of direction-selective neurons in LIP
        from state 1 to state 2,
        thereby increasing both the differences in input currents
        between the two groups and their absolute magnitudes.
        % \item
        \textbf{b}.
        Changes in the information received by LIP affect the
        energy landscape and attractors in the decision space.
        %  Weak
        Insufficient input currents fail to trigger a decision,
        maintaining LIP neurons at state 1;
        optimal input currents and current differences
        drive the state transition to decision state 2;
        excessively large input currents cause
        the state to shift towards 3,
        deviating from the intended decision state.
        % \item
        \textbf{c}.
        Changes in LIP's recurrent connections also alter
        the landscape and attractor states in the decision space.
        Insufficient recurrent connections result in
        weak evidence being unable to drive LIP
        from resting state 1 to decision state 2;
        excessive recurrent connections cause
        the resting attractor state 1 to disappear
        and enlarge the attraction basins of decision states 2 and 3,
        where weak inputs or noise can drive LIP to one of these decision states.
    % \end{enumerate*}
    }
\end{figure}

In the model, LIP Neurons primarily rely on
differences in input currents for decision-making.
Larger input currents enable the model to reach
the decision threshold more quickly but reduce accuracy.
For instance, increasing the connection strength
between V1 and MT, between MT and LIP,
or increasing the number of MT neurons connected to LIP
enhances the total currents received by LIP neurons
(see Fig~\ref{fig:diagram}a).
This enhancement improves signal differentiation
among neuronal groups that prefer different directions,
thereby increasing accuracy (see Fig~\ref{fig:tuning}b\&d, first half).
However, larger input currents cause attractors in decision space
to shift towards the diagonal
(see Fig.~\ref{fig:diagram}b, from 2 to 3) \cite{ye2021quantifying}.
In this scenario, inhibitory neurons are less effective
at suppressing DS neurons that prefer the opposite direction,
disrupting the \emph{winner-take-all} dynamics
and deteriorating model performance
(see Figs.~\ref{fig:tuning}b\&d, second half).

Similarly, modifying the recurrent connection weights
among LIP neuronal groups alters the energy landscape of the decision space.
Stronger recurrent connections eliminate the resting states attractor
and expand the size of the decision state attractors
(see Fig.~\ref{fig:diagram}c) \cite{ye2021quantifying}.
This modification improves performance during the initial phase
(see Fig.~\ref{fig:tuning}h, first half),
but excessively large decision attractors
can lead to rapid decision-making in the presence of noise,
thereby reducing accuracy (see Fig.~\ref{fig:tuning}h, second half).
Adjusting the number of LIP neurons also
alters the energy characteristics in the decision space.
If the parameters result in stable decision-state attractors,
the model can make decisions and maintain the states without external stimuli,
demonstrating a degree of working memory capability.
This enables decision-making with limited information
by integrating noisy evidence,
which can shorten decision time but introduces a risk of errors.
Conversely, if only one resting state attractor exists,
the model fails to reach the decision threshold
under any stimulus and cannot make decisions,
as evidenced by the perturbation experiment (Fig.~\ref{fig:noise}h, blue),
where we deliberately chose the preferred direction
associated with higher firing rate neurons for decision.
In real-world scenarios,
the model would not make a choice in this situation.
Similarly, parameter adjustments can lead to another monostable mode,
where neuron groups preferring opposite directions
both exhibit high firing rates,
resulting in an inability to make reasonable decisions
(Figs.~\ref{fig:noise}g\&i, purple).

In 2-alternative forced choice (2-AFC) tasks,
researchers often model decision-making as evidence accumulation over time,
using approaches such as
the drift diffusion model (DDM)
\cite{ratcliff1978theory, ratcliff2006modeling, ratcliff2008diffusion},
racing diffusion model \cite{usher2001time, tillman2020sequential},
and linear ballistic accumulator (LBA) model \cite{brown2008simplest}.
The temporal integration in the LIP module resembles the DDM
(see discussions in
\cite{wang2002probabilistic, wong2006recurrent, wang2008decision, wei2017decision}%
), and can be directly related to it
\cite{roxin2008neurobiological, umakantha2022relating}.
This type of evidence accumulation is not present in feedforward DCNN models.
Additionally, the dynamic nature of the LIP constitutes an attractor network
\cite{wong2006recurrent},
where, through iterative processes,
the network's state stabilizes near attractors.
This differs fundamentally from traditional DCNNs,
which categorize through spatial partitioning within representational spaces,
and may explain why neural dynamics models
exhibit better stability and noise resistance.

In summary, integrating neural dynamics into AI models
can bridge the gap between biological intelligence and artificial systems,
leading to the development of next-generation AI
that is both interpretable and resilient like biological organisms.
By leveraging advances in neuroscience to construct models with neural dynamics,
we are able to create AI systems that better mimic biological behavior.
The neuroimaging-informed fine-tuning method enhances performance
while preserving the advantages of neural dynamics.
This approach ensures both high performance and explainability,
aligning with biological plausibility and computational efficiency.
This synthesis between advanced AI techniques
and in-depth neuroscientific insights represents a promising direction
for future research and development in both fields.

%%%%%%%%%%%%%%%%%%%%%%%%%%%%%%%%%%%%%%%%%%%%%%%%%
\section{Methods}
\label{sec:methods}

\subsection{The RDK Dataset}
\label{method:rdk}

Random Dot Kinematogram (RDK)
is a classic and common psychophysical paradigm
used to study visual motion perception.
In a typical RDK stimulus,
numerous small dots are randomly distributed on the screen
moving at different speeds in different directions,
without a clear pattern.
This randomness is crucial,
preventing subjects from relying solely on local motion cues
to judge overall motion characteristics.
Typically, a certain proportion of dots (signal dots)
move in a specified direction (target direction),
while the remaining dots (noise dots)
move randomly \cite{kim1999neural, roitman2002response}.
Movement in a fixed direction is referred to as coherent motion,
and observers' ability to report the direction of coherent motion
increases with the percentage of coherent motion dots,
accompanied by shorter reaction times.
This proportion known as coherence,
is a measure of task difficulty
\cite{britten1992analysis},
making it the most important control parameter in RDK stimuli.
RDK can conveniently control the relative saliency of motion stimuli,
making it suitable for coherence threshold detection
or measuring changes in subject behavior with coherence.

In a 2-alternative forced choice (2-AFC) task,
with the proportion of coherent motion or
its logarithm as the independent variable,
a psychometric curve can be plotted
showing the probability of a subject choosing a certain direction.
This curve typically follows a sigmoid shape and can be described
by the equation (\ref{eq:psychometric}).
\begin{equation}
    p = \frac{1}{1+e^{-kx+b}}.
    \label{eq:psychometric}
\end{equation}
Here, the slope $k$ of the linear part
is referred to as \emph{sensitivity},
with higher sensitivity indicating a better ability
to distinguish between two directions of motion
and better behavioral performance.
The intercept $b$ describes the bias of the decision-maker,
i.e., the extent to which the decision-maker
tends to choose one direction over the other.
A good decision-maker should have higher sensitivity and smaller bias.

To ensure consistency in data during model testing,
we generated different RDK stimuli using
the specialized neuroscience and experimental psychology software
PsychoPy \cite{peirce2019psychopy2}
with varying parameters and stored the stimuli
in the form of three-dimensional arrays.

Computer vision systems that use a camera as the perceptual input
typically measure input videos and images in pixels,
rather than in degrees as in studies of biological vision.
Therefore, the RDK animation used in this study is pixel-based.
Specifically, within a rectangular black background
with dimensions of 300 pixels by 300 pixels,
a 270-pixel diameter aperture is placed at the center.
Inside the aperture,
200 white circles with a diameter of 6 pixels are evenly distributed.
Some points (signal dots) are randomly selected to move left or right,
while the remaining points (noise dots) are randomly dispersed in all directions.
The circles move at a speed of 2 pixels per frame,
and their positions are randomly reset every 4 frames
or when they move out of the aperture.
Following these rules,
120 frames of images are continuously generated
and stored in 8-bit grayscale format
as a matrix of size \numproduct{300x300x120}.
This animation lasts for 2 seconds on a \qty{60}{\Hz} monitor,
matching the stimulus duration of the human behavioral experiment.
The RDK dataset contains 100 levels of coherence
ranging from \qtyrange{0}{99}{\percent}
with a step size of \qty{1}{\percent}.
For each coherence level,
10 stimuli were generated for both leftward and rightward motion directions,
resulting in a total of \num{2000} stimuli.
This fine-grained variation in coherence levels allows for a detailed analysis
of the model's sensitivity to subtle differences in motion direction.

\subsection{The Neural Dynamics Model}
\label{method:model}

The neural dynamics model constructed in this study
simulates four key areas of the dorsal pathway
involved in motion perception,
corresponding to the LGN, V1, MT and LIP regions
of the motion perception decision system
from primary to high levels (Fig.~\ref{fig:model}a).
We did not directly model lower visual pathway modules
(retina, ganglion cells, etc.) in detail,
but simplified them into temporal and spatial convolution calculations,
directly mapping visual input to the total input current
of LGN neurons as the model's input.

The LGN layer of the model was constructed
based on previous research
\cite{chariker2021theory,chariker2022computational}.
The LGN is divided into two groups of \emph{ON} and \emph{OFF} neurons
adapting the Leaky Integrate-and-Fire (LIF) model,
each group containing \num{10000} neurons,
all with double Gaussian spatial receptive fields
(Equation~\ref{eq:spatial}, Fig.~\ref{fig:model}b middle).
These neurons cover a visual angle of \qty{0.35}{\degree}
($x, y \in [-0.175, 0.175]$ in Equation~\ref{eq:spatial}),
corresponding to a \numproduct{9x9} pixel area.
They are alternately overlapped and arranged regularly in space
(Fig.~\ref{fig:model}b top),
covering the entire \numproduct{300x300} field of view,
forming a two-dimensional plane corresponding to the image space.
\emph{ON} and \emph{OFF} neurons have different temporal profiles,
with \emph{ON} neurons responding slower than \emph{OFF} neurons
(approximately \qty{10}{\ms},
Equation~\ref{eq:temporal}, Fig.~\ref{fig:model}b bottom).
Images of \numproduct{300x300} pixels are convolved spatially
with the \numproduct{9x9} spatial convolution kernels,
then the time step is reduced to \qty{2}{\ms}
(120 frames expanded to 1000 frames)
through nearest-neighbor interpolation,
and temporal convolution with a sliding time window of \qty{160}{\ms} is performed.
At each time point,
two sets of \numproduct{100x100} convolution results are obtained,
which serve as the stimulus-induced current
for each neuron in the \emph{ON} and \emph{OFF} neuron groups, respectively,
and together with Ornstein--Uhlenbeck noise
\cite{lansky2001ornstein}
form the input current of LGN neurons.
\begin{equation}
    A(x, y) =
    \frac{\alpha}{\pi \sigma_\alpha^2} \exp\left(
        -\frac{x^2+y^2}{\sigma_\alpha^2}
    \right)
    - \frac{\beta}{\pi \sigma_\beta^2} \exp\left(
        -\frac{x^2+y^2}{\sigma_\beta^2}
    \right)
    \label{eq:spatial}
\end{equation}
where $\alpha = 1$, $\beta = 1$,
$\sigma_\alpha = 0.0894$,
$\sigma_\beta = 0.1259$
\cite{chariker2021theory}.
\begin{equation}
    K(t) =
    \alpha \frac{t^6}{\tau_0^7} \exp\left(
        -\frac{t}{\tau_0}
    \right)
    - \beta \frac{t^6}{\tau_1^7} \exp\left(
        -\frac{t}{\tau_1}
    \right)
    \label{eq:temporal}
\end{equation}
where $\tau_0 = 3.66$, $\tau_1 = 7.16$,
$\alpha = 1$, $\beta = 0.8$ for \emph{ON} neurons, and
$\alpha = 1$, $\beta = 1$ for \emph{OFF} neurons
\cite{chariker2021theory}.

% \begin{equation}
%     % d x_t = \theta(\mu - x_t) dt + \sigma d W_t
%     d x_t = \theta x_t dt + \sigma d W_t
%     \label{eq:ou}
% \end{equation}
% where $\sigma = 5$.

The other three brain regions are composed of LIF neuron groups,
with each neuron group connected to the neurons
in the previous layer with specific structures and probabilities.
The V1 area has a total of 5000 neurons,
divided into two groups, namely \emph{G1} and \emph{G2}.
Each V1 neuron receives input from
one \emph{ON} cell and one \emph{OFF} cell
arranged in a specific pattern
through AMPA synapses.
In the LGN projection received by the \emph{G1} group neurons,
the \emph{ON} cells are always located to the left of the \emph{OFF} cells.
Similarly, in the LGN projection received by the \emph{G2} group neurons,
the \emph{ON} cells are always located to the right of the \emph{OFF} cells
(Fig.~\ref{fig:model}c).
The MT area contains two groups of neurons, \emph{L} and \emph{R},
each consisting of 400 neurons.
The \emph{L} group neurons receive input from the \emph{G1} group neurons
within a certain receptive field,
while the \emph{R} group neurons receive input from
the \emph{G2} group neurons within the same receptive field.
The connections between MT and V1 corresponding neuron groups
are also formed by AMPA synapses,
with a synaptic conductance of
$\bar{g} = \qty{2.0}{\nano\siemens}$.

The LIP area is constructed according to the model
in literature \cite{wang2002probabilistic},
consisting of excitatory neuron groups \emph{A} and \emph{B} (each with 300 neurons)
and an inhibitory neuron group \emph{I} (500 neurons).
The \emph{A} group neurons receive random projections from \emph{L},
while the \emph{B} group neurons receive projections from \emph{R}
(\qty{50}{\percent} of neurons for each group) with AMPA synapses.
Connections within excitatory groups \emph{A}, \emph{B},
as well as connections from excitatory groups to inhibitory neuron group \emph{I},
are formed by AMPA and NMDA synapses with different temporal characteristics.
The inhibitory neuron group \emph{I} inhibits neurons
in groups \emph{A} and \emph{B} through GABA synapses.
The connection strengths (synaptic conductance coefficients)
of the above connections follow a normal distribution
$\mathcal{N}(\bar{g}, 0.5\bar{g})$,
where the average conductance from MT to LIP is
$\bar{g}_{MT} = \qty{0.1}{\nano\siemens}$,
the average conductance among excitatory neurons in LIP is
$\bar{g}_{AMPA} = \qty{0.05}{\nano\siemens}$,
$\bar{g}_{NMDA} = \qty{0.165}{\nano\siemens}$,
the average conductance from excitatory neurons to inhibitory neurons in LIP is
$\bar{g}_{AMPA} = \qty{0.04}{\nano\siemens}$,
$\bar{g}_{NMDA} = \qty{0.13}{\nano\siemens}$.
If the connection strength between two neurons is less than 0,
there is no connection (Fig.~\ref{fig:model}c).
Additionally, the connection weights between neurons
with the same direction preference increase to
$w = 1.3$ times the original weight (\emph{Hebb-strengthened weight}),
while the connection weights between neurons
with different direction preferences weaken to
$w = 0.7$ times the original weight (\emph{Hebb-weakened weight}).

The neurons in the model are all LIF neurons,
a commonly used model in computational neuroscience
to approximate the behavior of biological neurons.
The resting membrane potential is set to $V_r = \qty{-70}{\mV}$,
and when the membrane potential of a neuron reaches \qty{-50}{\mV},
it fires an action potential,
resetting to \qty{-55}{\mV}
during the refractory period \cite{troyer1997physiological}.
The LIF model was selected due to its
simplicity and computational efficiency,
while still capturing essential dynamics of neuronal firing.
In addition to synaptic currents and external currents,
each neuron receives Ornstein--Uhlenbeck noise,
a standard approach for modeling stochastic fluctuations in neuronal input.
The time constant of OU noise $\tau = \qty{10}{\ms}$,
and mean current of \qty{400}{\pA} with a variance of \num{100}
were chosen to match the noise characteristics observed in biological neurons.
for LIP excitatory neurons,
a higher mean current of \qty{550}{\pA} was used to simulate
the enhanced excitatory drive from
other excitatory neurons without direction selectivity.
For excitatory neurons, the parameters are
$C_m = \qty{0.5}{\nF}$,
% $C_m = \qty{500}{pF}$,
$g_l = \qty{25}{\nano \siemens}$,
and refractory period $\tau = \qty{2}{\ms}$.
For inhibitory neurons, the parameters are $C_m = \qty{0.2}{\nF}$,
$g_l = \qty{20}{\nano \siemens}$,
and refractory period $\tau = \qty{1}{\ms}$.
The AMPA, NMDA, and GABA synapses in the model
follow the settings in the literature
\cite{wang2002probabilistic, jahr1990voltage}.

We use the overall firing rate of neuron groups
\emph{A} or \emph{B} in the LIP region
as the basis for whether the model makes a decision.
When the average firing rate exceeds a threshold (\qty{30}{\Hz})
or when the stimulus finishes,
the model selects the direction preferred by
the neuron group with the higher firing rate as its final output.
By simulating the model with each stimulus in
the aforementioned RDK dataset,
we obtain the accuracy of the model's decisions
and the number of time steps taken for the decision (decision time)
under different coherence levels.
We can estimate a psychometric function through
least squares regression (Equation~\ref{eq:psychometric}).
This method yields behavioral metrics,
including the model's psychometric curve and sensitivity indices.

Furthermore, our model,
which possesses a structure akin to biological systems
and neurons that mirror those of the biological nervous system,
allows for the execution of virtual electrophysiological experiments,
recording and analyzing the firing characteristics
of each neuron group in the model,
and even performing virtual electrical stimulation on neurons
by adding additional current inputs,
to study the characteristics of the model
(Figs.~\ref{fig:behavior}b\&c).

Due to the randomness of RDK stimuli and the neural dynamics of the model,
to ensure the robustness of the results,
each stimulus was repeated twice in the model performance test,
and the neural dynamics model was re-initialized and repeated five times.
The selection probabilities and average decision times
for each coherence level were calculated and
estimated using the psychometric curve (Equation~\ref{eq:psychometric}),
and the decision time curve was estimated using
a moving median smoothing algorithm with a coherence window of \qty{10}{\percent}
(Fig.~\ref{fig:behavior}a upper panel).
For human subjects,
we conducted a bootstrap analysis on the behavioral data,
where the trials of each subject were randomly sampled with replacement,
and the median of 1000 bootstrap samples was calculated for each subject.
The final average performance over all subjects was plotted.

\subsection{A CNN Model for Motion Perception}
\label{method:motionnet}

A convolutional neural network (CNN) model (\emph{MotionNet}) was built
similar in structure and scale to the neural dynamics model for comparison
(Fig.~\ref{fig:noise}a).
The model receives video input with a resolution of \numproduct{300x300} pixels,
processes it through spatial convolutions of \numproduct{9x9} pixels
and temporal convolutions of 10 frames,
resulting in two \numproduct{100x100} feature maps
matching the input of LGN neurons.
These feature maps are converted to two \numproduct{50x50} maps
using \numproduct{3x3} convolutional kernels.
The two channels are then processed with
\numproduct{11x11} convolutional kernels to generate two \numproduct{20x20} feature maps.
After average pooling along the time dimension,
these features are mapped to a 400-dimensional vector by a linear layer,
and finally to output neurons using another linear layer.
All neurons utilize the \emph{ReLU} activation function except the output layer,
which is passed through a softmax transformation
for fitting the one-hot encoded direction classification information.

To avoid overfitting the training data and ensure accurate model comparison,
\emph{MotionNet} was not directly trained on the RDK dataset.
Instead, it utilized generated moving random images as the training data.
Specifically, a random grayscale image was generated and stretched randomly.
A window covering the image was moved in random directions
to obtain the necessary animation data.
The speed and direction of movement were randomly assigned,
and the training labels (supervision information, moving left or right)
were determined by the direction of movement along the horizontal axis.

\emph{MotionNet} was trained with the cross-entropy loss function
and stochastic gradient descent (SGD) method.
The initial learning rate was set at \num{0.01},
reduced to \qty{10}{\percent} at the 5th and 15th epochs.
The momentum was set to \num{0.9},
with a batch size of \num{64},
and each epoch contained 500 batches.
Training stopped at the 20th epoch.
The model from the 20th epoch was selected for testing
based on its convergence and stability during training.
Similar analyses were then conducted on this model using the RDK dataset,
paralleling those applied to the neural dynamics model.
This approach ensured a fair comparison between
the two models' performance under the same experimental conditions.
Specifically, twenty consecutive frames were randomly selected from
a 120-frame animation as input for \emph{MotionNet}.
The model's choices under various coherences were recorded
and psychometric curves were fitted.
For result stability, each trial was repeated twice.
Unlike the neural dynamics model, \emph{MotionNet},
being a CNN, lacks a concept of time,
so we focused on its psychometric curve
and related parameters (\emph{sensitivity}).

\subsection{RDK Behavioral and Neuroimaging Experiments in Human Subjects}
\label{method:subjects}

Subjects:
Thirty-six subjects participated in this RDK experiment
(12 males, 24 females, mean age 29 \textpm 8 years).
Each subject conducted a behavioral experiment
with a series of random-dot motion-direction discrimination tasks
(consisting of a learning period, a practice period, and a test period)
and an MRI scanning experiment.
The MRI experiment consisted of
a high-resolution 3D-T1 structural scanning session,
a field-map scanning session for image distortion correction,
a functional-MRI session for localization of the MT region,
two functional-MRI sessions conducting the RDK tasks,
and a DWI session for white-matter fiber reconstruction.
The study was approved by the Ethics Committee,
and informed consent was obtained from each volunteer.

Behavioral experiment:
Random-dot stimuli were presented on a monitor
at a distance of \qty{57}{\cm} from the subject,
with a display resolution of \numproduct{2560x1440} pixels
and a refresh rate of \qty{60}{\Hz}.
The stimulus program was written using PsychoPy
\cite{peirce2019psychopy2}.
The stimuli were a series of circular dots
(white dots presented on a black background).
Each trial was initialized with a fixation cross
(\qtyproduct{.33 x 0.33}{\degree} dva,
lasting for a fixed duration of \qty{500}{\ms}
plus a random duration sampled from a uniform distribution
with a maximum of \qty{200}{\ms}).
Subjects were asked to gaze at the cross.
After that, a set of moving dots was presented
in a circular area of 5 dva diameter,
with a total number of 300 dots and
a diameter of \ang{0.04} dva for each of these dots.
Some of the dots moved uniformly to the left or right,
while the rest of the dots moved randomly with a speed of
% 3.3 degrees per second.
\qty{3.3}{\degree\per\s}.
The positions of all the dots were re-randomized every 5 frames
(i.e., the presentation of each dot lasted 5 frames).
Subjects were asked to recognize the uniform direction
of the moving dots and respond within
\qty{2000}{\ms} (Extended Fig.~\ref{fig:paradigm}a).
Each subject first performed a learning period.
The proportion of the uniformly moving dots for this period
was fixed at a coherence of \qty{80}{\percent}
(i.e., \qty{80}{\percent} of the dots move in the same direction),
and the subject was informed with correct or incorrect feedback
after responding to the direction.
This period would not terminate
until the subject responded correctly 10 consecutive times.
After the learning period, the subject entered a practice period.
A total of 5 sessions were presented
and each session consisted of 70 trials
at randomized coherence levels
(\qtylist{1;2;3;4;5;6;8;10;15;20;25;30;40;50}{\percent},
for a total of 14 coherence levels, each of which appeared 5 times).
During this period,
subjects were also informed of correct or incorrect feedback after responding.
Finally, the subject would be presented with a staircase test.
The coherence of the trials was set by the staircase method (3-down 1-up method),
with an initial coherence of \qty{10}{\percent},
a step size of \qty{5}{\percent} before the first reversal,
and a step size of \qty{1}{\percent} after the first reversal.
The behavioral experiment was completed after the 20th reversal or 150 trials in total.
A psychometric curve was then fitted for each subject's performance,
and a coherence corresponding to \qty{79.4}{\percent} accuracy
was estimated as the initial value of coherence in the following MRI experiment.

MRI experiment:
MRI experiments were performed using a MAGNETOM Prisma 3T
(Siemens Healthcare, Erlangen, Germany) MR scanner
with a 64-channel head-neck coil.
High-resolution 3D T1 structural images were acquired
using the MPRAGE sequence with the following scanning parameters:
% TR=2530 ms, TE=3.34 ms, TI=1000 ms, flip angle=7°, iPAT=2, FOV=256*256 mm2,
% number of slices=192, resolution=1 mm isotropic.
TR=\qty{2530}{\ms},
TE=\qty{3.34}{\ms},
TI=\qty{1000}{\ms},
flip angle=\ang{7},
iPAT=2,
FOV=\qtyproduct[product-units=power]{256x256}{\mm},
number of slices=192,
resolution=\qty{1}{\mm} isotropic.
Field-map images were scanned using a dual-echo 2D-GRE sequence
with the following parameters:
% TR=747 ms, TE1=4.92 ms, TE2=7.38 ms, FOV=208*208 mm2,
% number of slices=72 slices, resolution=2.0 mm isotropic.
TR=\qty{747}{\ms},
TE1=\qty{4.92}{\ms},
TE2=\qty{7.38}{\ms},
FOV=\qtyproduct[product-units=power]{208x208}{\mm},
number of slices=72 slices,
resolution=\qty{2.0}{\mm} isotropic.
Functional-MRI was performed using a 2D-GRE-EPI sequence
with simultaneous multi-slices scanning with the following parameters:
% TR=1000 ms, TE=35 ms, flip angle=52°, FOV=208*208mm2,
% number of slices=72 slices, resolution=2.0 mm isotropic, SMS factor=8.
TR=\qty{1000}{\ms},
TE=\qty{35}{\ms},
flip angle=\ang{52},
FOV=\qtyproduct[product-units=power]{208x208}{\mm},
number of slices=72 slices,
resolution=\qty{2.0}{\mm} isotropic,
SMS factor=8.
The diffusion-weighted images were scanned using a 2D-SE-EPI sequence
with simultaneous multi-slices. Scanning parameters:
multiple b-values (b=0, 1000, 2500),
96 diffusion directions (32 b=1000 volumes and
64 b=2500 volumes \cite{tian2022comprehensive})
and 9 b=0 volumes,
TR=\qty{7400}{\ms},
TE=\qty{70}{\ms},
iPAT=2,
FOV=\qtyproduct[product-units=power]{204x204}{\mm},
number of slices=96 slices,
resolution=\qty{1.4}{\mm} isotropic,
SMS factor=2,
partial fourier=6/8,
phase-encoding direction=A$>>$P with only one b=0 volume P$<<$A.

The MT Localizer task used the classic random-dot
expansion-contraction paradigm
\cite{michels2005visual}.
The stimuli were presented on a monitor
at a distance of \qty{170}{\cm} from the subject,
with a resolution of \numproduct{1920x1080} pixels
and a refresh rate of \qty{60}{\Hz}.
The program was written using PsychoPy.
The stimuli were a series of white circular dots presented on a black background.
The experimental paradigm was designed to be an \verb|rArArA| block design,
consisting of a \qty{16}{\s} rest block
alternating with a \qty{16}{\s} visual stimuli block
lasting for a total duration of \qty{112}{\s}.
A fixation cross was first presented
at the center of the screen for \qty{16}{\s},
and subjects were asked to gaze at the cross;
three blocks were then presented.
Each block consisted of 200 dots (\ang{0.15} dva)
presented in a ring ranging from \ang{0.5} dva to \ang{9} dva.
A bigger dot of \ang{0.3} dva was presented at the center of the screen,
and then all white dots except the central dot
started a contraction-expansion movement
with a speed of \qty{8}{\degree\per\s}.
Subjects were asked to look at the central dot throughout the entire process.

The RDK test task in the MRI scanning used
the same paradigm as the behavioral test.
Subjects were asked to press two buttons with either the left or right hand.
The task adopted an \verb|rABrAB| block design,
consisting of a % 4-second rest block,
\qty{4}{\s} rest block,
a \qty{4}{\s} random viewing block,
and a \qty{4}{\s} prompt-and-press block.
Each subject went through two sessions,
each consisting of 25 trials,
and a total duration of \qty{304}{\s}.
The experimental stimuli of each trial
were consistent with those presented in the behavioral experiment,
except that in the prompt-and-press block,
a new cue of `left-handed' or `right-handed' appeared
at the center of the screen
and the subjects followed the cue
to press the key with the corresponding hand
(Extended Fig.~\ref{fig:paradigm}b).

In the MRI experiment,
we found that the RDK stimulus-induced activation was
mainly located in the primary visual cortex, MT, and LIP regions
(Extended Fig.~\ref{fig:paradigm}c,
$p<0.001$, uncorrected, one-sample $t$-test).
This result was consistent with the ROIs modeled
in the neural dynamics model of motion perception.

\subsection{Neuroimaging Data Analysis Methods}
\label{method:imaging}

High-resolution 3D--T1 structural images were
parcellated into gray matter and white matter
and reconstructed into cortex surfaces using Freesurfer
\cite{fischl2012freesurfer}.
A voxel-level segmentation of brain regions
based on the \emph{Destrieux} atlas template
\cite{destrieux2010automatic} was also obtained.
Several indices were estimated,
including the number of voxels per subregion,
cortical area, and average cortical thickness.
These processed data and structural features were used
in the subsequent correlation analysis.

The fMRI images were corrected using FSL,
including slice timing correction,
motion correction, and field-map correction.
The corrected images were then aligned to
the individual space of the T1 structural image.
The WM/GM/CSF binary maps were constructed
in the structural space for the subsequent GLM analysis
to remove white matter and cerebrospinal fluid voxels.
In the first-level analysis, the GLM model was used,
and the rest, stimuli, and response blocks of each trial
were considered as separate regressors.
The six motion parameters
(three translation and three rotation parameters per frame)
and the white matter and cerebrospinal fluid temporal signals
were used as nuisance regressors in the model.
The GLM model was fitted to each voxel,
and the contrast of `stimuli $-$ rest' was constructed
to obtain the activation map in $z$-score.
In the second-level analysis, each subject's activation map in $z$-score
was projected to the individual cortical space
(reconstructed cortical surface),
and then projected to a standard cortical space (\emph{fsaverage5} template),
where the averaged activation map in $z$-score
was obtained by performing a one-sample $t$-test
($p<0.001$, uncorrected, Extended Fig.~\ref{fig:paradigm}c).

The resting-state functional connectivity in this study
was obtained from the task-fMRI data \cite{fox2016combining}.
By regressing out the task-dependent signals
and filtering the preprocessed results in the
\qtyrange{0.01}{0.08}{\Hz} band,
the background signals of the brain were obtained.
The resting-state functional connectivity matrix
was then calculated using two atlases:
the \emph{Destrieux} structural atlas
and the \emph{Glasser} functional atlas \cite{glasser2016multi}.
Each element of the matrix was used to characterize each ROI pair
to correlate with the behavioral results.

Data preprocessing steps for diffusion images included
brain extraction, registration to T1 images,
field-map top-up correction, and eddy-current correction.
We applied a ball-and-stick model to regress the white matter voxels.
The model regression of voxels in the white-matter region
inferred the white-matter orientation distribution of single voxels,
and FA and MD values for each voxel were then obtained
\cite{hamalainen2017bilingualism}.
By registering to the white-matter segmentation image,
white-matter regions' mean FA and mean MD values
(\emph{Destrieux} atlas, threshold: FA\num{>0.2})
were calculated for the subsequent correlation analysis.
Mean FA and MD values were used to characterize
the structural connectivity of white-matter pathways
to correlate with the behavioral results.

To visualize the fiber bundles of the subjects,
fiber tract reconstruction was performed at the individual level
using the quantitative anisotropy of
the \emph{DSI Studio}'s gqi inference as a tracking index.
Segmented ROIs (lateral occipital and inferior parietal regions)
were used as spatial constraints for fiber tracking.
Deterministic fiber tracking was performed by the streamline Euler method
(tracking threshold \num{=0},
angular threshold \num{=0},
min length \qty{=30}{\mm},
max length \qty{=300}{\mm},
seed number \num{=100000})
to obtain the morphology of fiber bundles
related to the visual decision-making brain regions
in the occipital and parietal lobes of the subjects.

\subsection{Neuroimaging-Informed Fine-Tuning}
\label{method:tuning}

Data analysis:
The performance of the subjects in the behavioral RDK experiment
was quantitatively analyzed using psychometric curves,
and Pearson correlation analysis was performed
between the behavioral results and the structural and functional features
estimated from the MRI data.
For behavioral data,
a psychometric curve was fitted to each subject's staircase performance,
and perceptual ability was determined by taking
% `1 - 79.4% accuracy'.
`1 $-$ \qty{79.4}{\percent} accuracy'.

Pearson correlations were calculated between the above features from structural
(number of voxels per subregion, cortical area, average cortical thickness),
diffusion (mean FA, mean MD),
and functional (resting-state FC elements) images
and the perceptual ability indices of the subjects.
The ROIs were determined by the parcellation
and registration to the \emph{Glasser} atlas
and the \emph{Destrieux} atlas in each individual space.
We calculated the correlations between the ROIs' features
and the behavioral perceptual abilities
($p<0.01$ as a threshold of statistical significance).

Connecting physiological data indicators found
in magnetic resonance imaging (MRI) with the model
is challenging due to the vastly different scales of neural elements involved.
In this study, a heuristic approach was used
to adjust and assess parameters potentially correlated with
structural and functional MRI indicators,
inspired by functional analogy.
White matter connections found in MRI were simulated
by adjusting the mean of the connection parameter distribution for V1 to MT
(\numrange{0.2}{3.0} with a step size of \num{0.2}),
or by adjusting the proportion of connections between MT and LIP
(\qtyrange{10}{100}{\percent} with step \qty{10}{\percent}).
According to \emph{Dale's principle},
non-positive connection weights were set to zero,
equivalent to removing those connections.
The adjusted model was retested on the RDK dataset five times,
and statistical analysis was performed on the sensitivity of
psychometric curves (\emph{slope})
to investigate the impact of parameter adjustments
on model performance and identify parameter combinations
that improve model performance.
Similarly, for functional connections found in fMRI,
the synaptic conductance of neurons in MT was altered
(\numrange{0.01}{0.15} with a step size of 0.01),
or the \emph{Hebb-strengthened weight}
between LIP excitatory neurons was changed
(\numrange{0.7}{1.5} with a step size of 0.1)
to simulate the neural modulation from other brain regions.
The same statistical analysis was conducted
to examine the effect of parameter adjustments on model performance.

\subsection{Interference Resistance Experiment of the Model}
\label{method:noise}

In addition to requiring large amounts of data for training,
deep learning models are also prone to biases
and are easily affected by noise,
resulting in poor transferability, among other issues.
To address these problems, we designed a series of perturbation experiments
on the neural dynamics model to assess its noise resistance performance.

For each module of the neural dynamics model,
we separately tested the effects of deactivating a certain percentage of neurons,
discarding a certain percentage of synapses
(ranging from \qtyrange{0}{90}{\percent} with a step size of \qty{10}{\percent}),
adding a certain level of Gaussian perturbation to the input current of neurons
(variance ranging from 0 to 2 times the
averaged absolute value of the group's input current,
normal distribution noise with a mean of 0),
and adding a certain level of Gaussian perturbation to the connection weights between neurons
(variance ranging from 0 to 2 times the
averaged absolute value of all connection weights,
normal distribution noise with a mean of 0).
For each parameter setting,
the model's behavioral performance was evaluated using the RDK dataset.
Our analysis concentrated on understanding
how changes in model parameters affected sensitivity and overall accuracy.
This evaluation highlights the significance of each parameter
and demonstrates the resiliency of our model.

For comparison, we conducted similar noise and damage experiments
on \emph{MotionNet} as we did on the neural dynamics model.
For each layer of \emph{MotionNet},
we separately tested the effects of
deactivating a certain percentage of neurons,
discarding a certain percentage of connections
(ranging from \qtyrange{0}{90}{\percent} with a step size of \qty{10}{\percent}),
adding a certain level of Gaussian perturbation to the output of neurons
(variance ranging from 0 to 2 times the
averaged absolute value of the group's neuron activation,
normal distribution noise with a mean of 0),
and adding a certain level of Gaussian perturbation to the connection weights between neurons
(variance ranging from 0 to 2 times the
averaged absolute value of all connection weights,
normal distribution noise with a mean of 0).
Due to the fact that the fully connected layer (fc4) of \emph{MotionNet}
is mapped to the output layer with only two neurons,
the fc4 connections were neither deactivated nor perturbed during our experiment.
After each adjustment of the perturbation parameters,
we conducted a complete evaluation on the RDK dataset
and calculated the sensitivity of the current model
to the parameter changes.

%%%%%%%%%%%%%%%%%%%%%%%%%%%%%%%%%%%%%%%%%%%%%%%%%
% tables are required to be at the end:
\begin{table}[htb]
    \centering
    \caption{\emph{slope of accuracy changes with respect to noise level}}
    \label{tab:compare}
\begin{tabular}{lcccc}
    \toprule
    noise type          & MotionNet & slope       & Neural Dynamics Model  & slope     \\
    \midrule
drop connection         & conv1    & \num{-0.530}  & LGN--V1       & \num{-0.458}  \\
drop connection         & conv2    & \num{-0.574}  & V1--MT        & \num{-0.081}  \\
drop connection         & fc3      & \num{-0.361}  & MT--LIP       & \num{-0.026}  \\
add noise to connection & conv1    & \num{-0.298}  & LGN--V1       & \num{-0.105}  \\
add noise to connection & conv2    & \num{-0.263}  & V1--MT        & \num{-0.270}  \\
add noise to connection & fc3      & \num{-0.041}  & MT--LIP       & \num{-0.006}  \\
drop neurons            & conv1    & \num{-0.594}  & LGN           & \num{-0.456}  \\
drop neurons            & conv2    & \num{-0.411}  & V1            & \num{-0.179}  \\
drop neurons            & fc3      & \num{-0.649}  & MT            & \num{-0.089}  \\
drop neurons            & fc4      & \num{-0.027}  & LIP           & \num{-0.476}  \\
add noise to neurons    & conv1    & \num{-0.004}  & LGN           & \num{-0.304}  \\
add noise to neurons    & conv2    & \num{-0.242}  & V1            & \num{-0.006}  \\
add noise to neurons    & fc3      & \num{-0.122}  & MT            & \num{-0.001}  \\
add noise to neurons    & fc4      & \num{-0.083}  & LIP           & \num{ 0.004}  \\
    \bottomrule
\end{tabular}
\end{table}

%%%%%%%%%%%%%%%%%%%%%%%%%%%%%%%%%%%%%%%%%%%%%%%%%
\clearpage
\backmatter
\section*{Extended Data}

\begin{appendices}
\renewcommand\theHfigure{sFigure.\arabic{figure}}
\renewcommand\theHtable{sTable.\arabic{table}}

\begin{figure}[htb]
    \includegraphics[width=\linewidth]{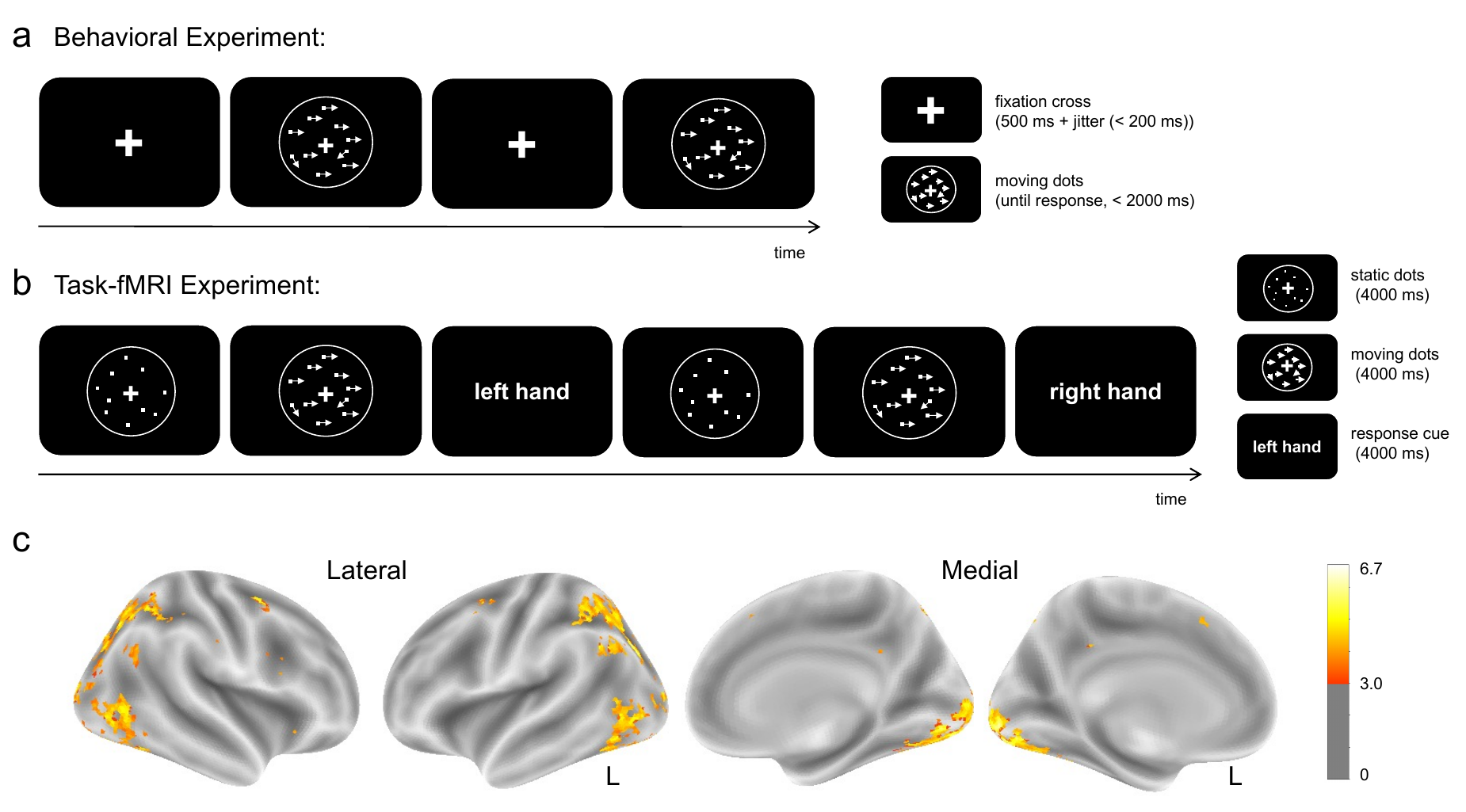}
    \caption[Experimental design of the behavioral and MRI tasks]{
        \label{fig:paradigm}
        \emph{Experimental design of the behavioral and MRI tasks.}
    % }
    % \justifying
    % \begin{enumerate*}[label=\textbf{\alph*.}]
        % \item
        \textbf{a}.
        The behavioral trial consisted of a fixation cross
        and a moving-dot stimulus for each trial.
        Subjects were instructed to fixate on the cross
        at the center of the screen throughout the entire task
        and to judge the global motion direction as quickly as possible
        (within \qty{2000}{\ms})
        during the stimulus presentation.
        % \item
        \textbf{b}.
        The MRI trial consisted of a fixation cross,
        a moving-dot stimulus, and a prompt stimulus.
        Subjects were asked to maintain fixation on the central cross
        and to press the corresponding button
        after the left-or-right prompt appeared.
        % \item
        \textbf{c}.
        Activation map ($z$-score) for the task,
        contrasting `stimulus' versus `rest'.
        Significant activation was observed in
        the primary visual cortex, MT, and LIP during the task
        (one-sample $t$-test, $p<0.001$, uncorrected).
    % \end{enumerate*}
    }
\end{figure}

\begin{figure}[htb]
    \includegraphics[width=\linewidth]{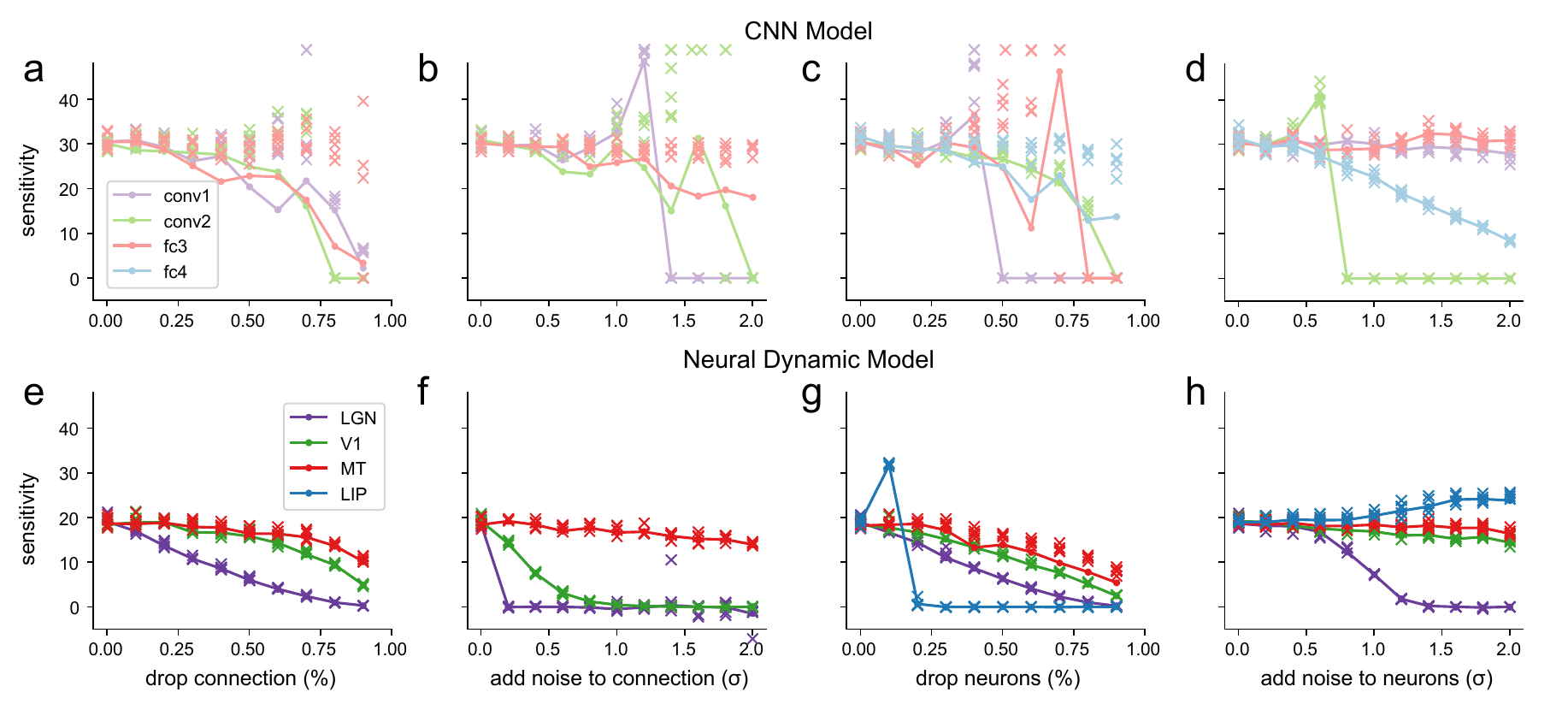}
    \caption[Sensitivity of CNN and the neural dynamics model changes with perturbation]{
        \label{fig:slope}
        \emph{Sensitivity of CNN and the neural dynamics model changes with perturbation.}
    % }
    % \justifying
    % \begin{enumerate*}[label=\textbf{\alph*.}]
        % \item
        \textbf{a}.
        Changes in model sensitivity
        when a certain percentage of connections
        are dropped in each layer of the CNN;
        \textbf{b}.
        Changes in model sensitivity
        when a certain level of Gaussian noise
        ($\sigma$ times averaged absolute value)
        are added to the connection weights of the CNN;
        \textbf{c}.
        Changes in model sensitivity
        when a certain percentage of neurons
        are dropped in each layer of the CNN;
        \textbf{d}.
        Changes in model sensitivity
        when a certain level of Gaussian noise
        ($\sigma$ times averaged absolute value)
        are added to the input of neurons of the CNN;
        % \item
        \textbf{e-h}.
        The same experiments for the neural dynamics model.
    % \end{enumerate*}
    }
\end{figure}

\begin{figure}[htb]
    \includegraphics[width=\linewidth]{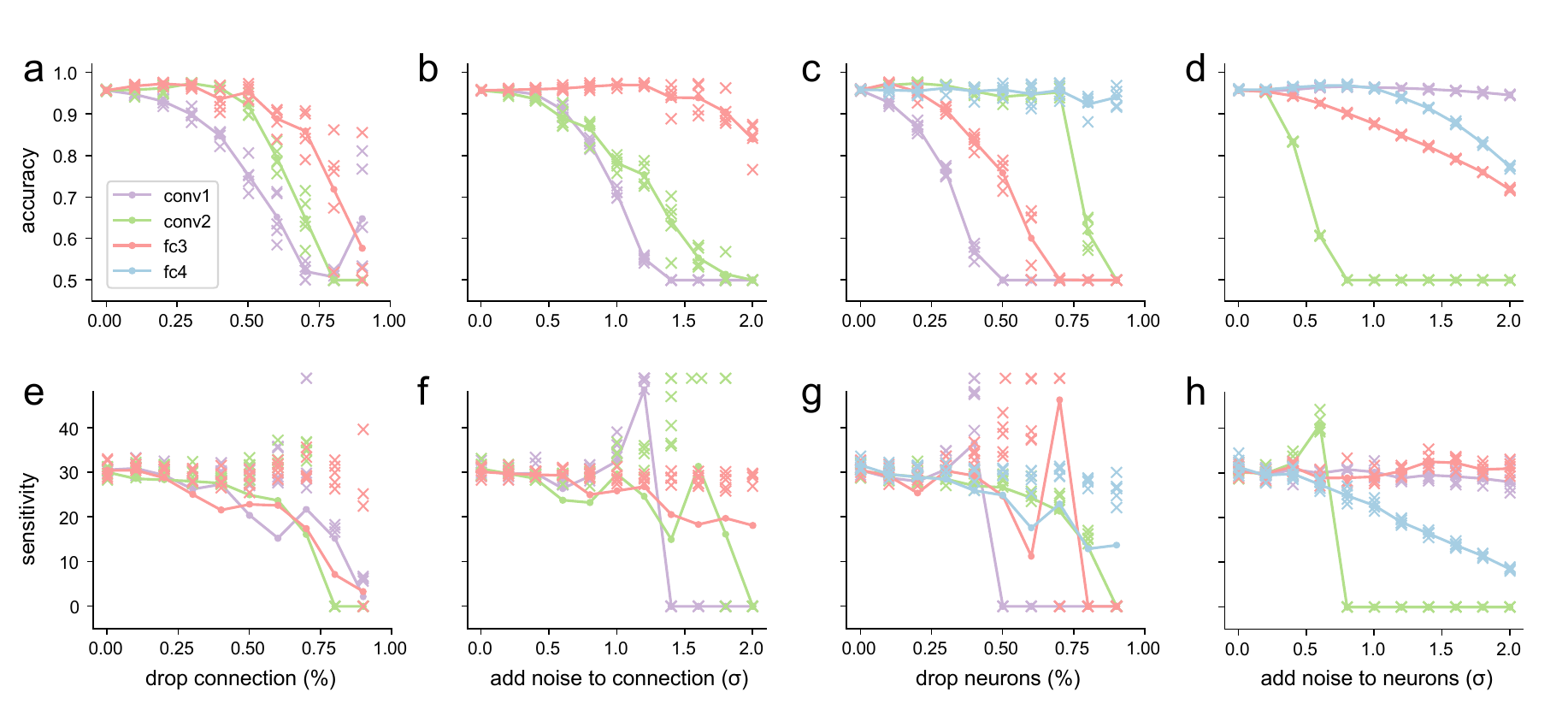}
    \caption[Perturbation experiment on another CNN model with the same structure]{
        \label{fig:another}
        \emph{Perturbation experiment on another CNN model with the same structure.}
    % }
    % \justifying
    % \begin{enumerate*}[label=\textbf{\alph*.}]
        % \item
        \textbf{a-d}.
        Accuracy of the model changes with noise level;
        % \item
        \textbf{e-h}.
        Sensitivity of the model changes with noise level.
    % \end{enumerate*}
    }
\end{figure}

\begin{figure}[htb]
    \includegraphics[width=\linewidth]{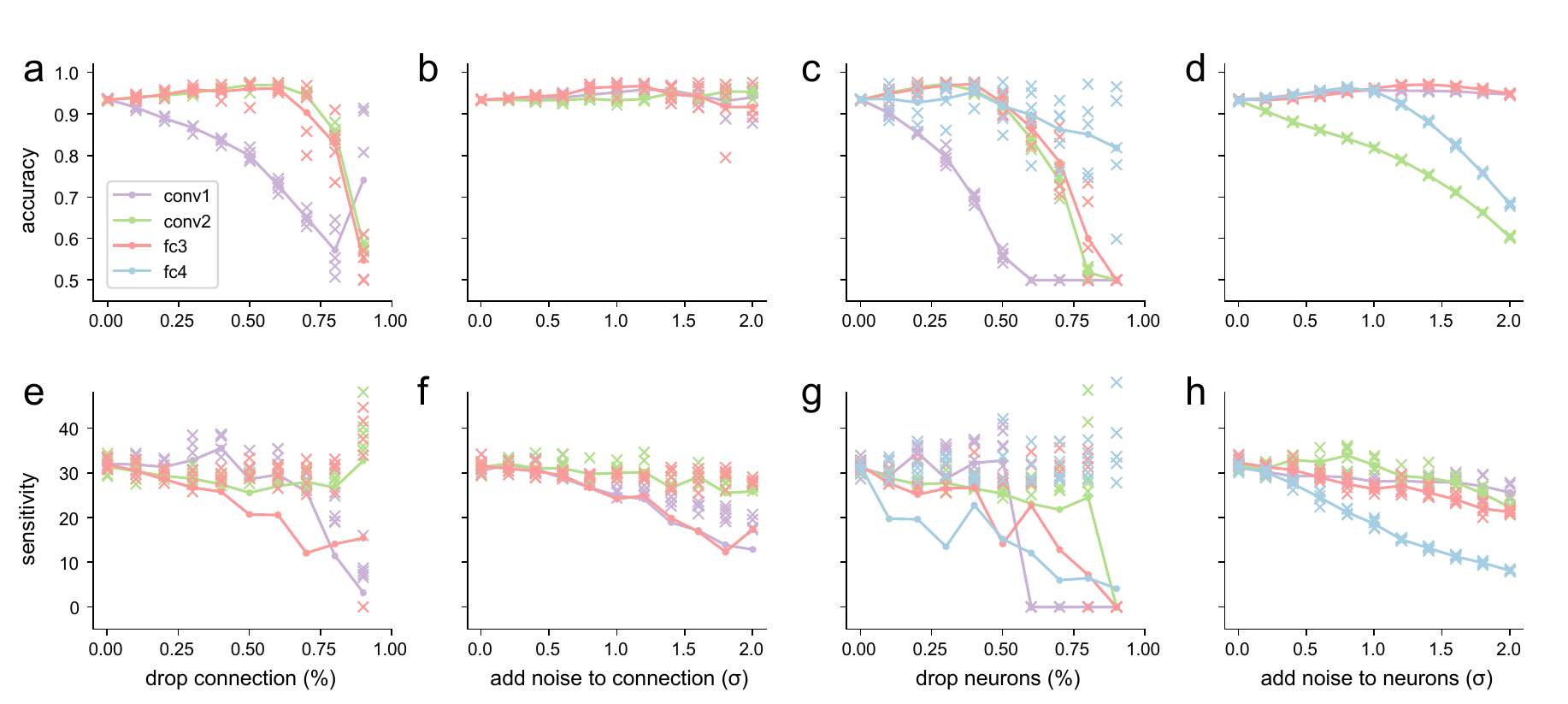}
    \caption[Perturbation experiment on another CNN model with the same structure]{
        \label{fig:onemore}
        \emph{Perturbation experiment on another CNN model with the same structure.}
    % }
    % \justifying
    % \begin{enumerate*}[label=\textbf{\alph*.}]
        % \item
        \textbf{a-d}.
        Accuracy of the model changes with noise level;
        % \item
        \textbf{e-h}.
        Sensitivity of the model changes with noise level.
    % \end{enumerate*}
    }
\end{figure}

\begin{table}[htb]
    \centering
    \caption{\emph{Slope of sensitivity changes with respect to noise level}}
    \label{tab:sensitivity}
\begin{tabular}{lcccc}
    \toprule
    noise type          & MotionNet & slope       & Neural Dynamics Model  & slope     \\
    \midrule
drop connection         & conv1    & \num{-17.4}  & LGN--V1       & \num{-22.4}  \\
drop connection         & conv2    & \num{-27.3}  & V1--MT        & \num{-14.3}  \\
drop connection         & fc3      & \num{-7.57}  & MT--LIP       & \num{-8.20}  \\
add noise to connection & conv1    & \num{-18.1}  & LGN--V1       & \num{-4.65}  \\
add noise to connection & conv2    & \num{-3.46}  & V1--MT        & \num{-8.39}  \\
add noise to connection & fc3      & \num{-0.96}  & MT--LIP       & \num{-2.48}  \\
drop neurons            & conv1    & \num{-46.9}  & LGN           & \num{-22.3}  \\
drop neurons            & conv2    & \num{-24.9}  & V1            & \num{-18.2}  \\
drop neurons            & fc3      & \num{-28.3}  & MT            & \num{-11.7}  \\
drop neurons            & fc4      & \num{-4.52}  & LIP           & \num{-24.3}  \\
add noise to neurons    & conv1    & \num{-1.10}  & LGN           & \num{-12.3}  \\
add noise to neurons    & conv2    & \num{-20.5}  & V1            & \num{-2.24}  \\
add noise to neurons    & fc3      & \num{ 0.87}  & MT            & \num{-0.93}  \\
add noise to neurons    & fc4      & \num{-12.0}  & LIP           & \num{ 3.02}  \\
    \bottomrule
\end{tabular}
\end{table}

\end{appendices}
\clearpage

%%%%%%%%%%%%%%%%%%%%%%%%%%%%%%%%%%%%%%%%%%%%%%%%%
% \bibliography{refs}
% \iffalse
% generated via bibtex with some modifications

% \fi

\end{document}